\documentclass[times,twocolumn,final]{elsarticle}

\usepackage{medima}
\usepackage{framed,multirow}
\usepackage{amsmath,amssymb,amsfonts}
\usepackage{latexsym}

\usepackage{url}
\usepackage{soul}
\usepackage{xcolor}
\usepackage{hyperref}
\usepackage{amsmath}

\usepackage{supertabular}
\usepackage{graphicx}
\usepackage{subfigure}
\usepackage{textcomp}
\usepackage{booktabs}
\usepackage{adjustbox}
\usepackage{multirow}
\usepackage{multicol}
\usepackage{makecell}
\usepackage{arydshln}
\usepackage{threeparttable}
\usepackage{longtable}

\usepackage{algorithmic}
\usepackage{algorithm}

\usepackage[normalem]{ulem}
\useunder{\uline}{\ul}{}

\colorlet{purple}{black}
\colorlet{blue}{black}

\usepackage[switch]{lineno}

\journal{Medical Image Analysis}
\begin{document}
\verso{T. Zhang \textit{et~al.}}

\begin{frontmatter}
\title{Self-Supervised Learning for Medical Image Data with Anatomy-Oriented Imaging Planes}

\author[1]{Tianwei \snm{Zhang}\fnref{fn1}}
\author[2]{Dong \snm{Wei}\fnref{fn1}}
\author[1]{Mengmeng \snm{Zhu}\fnref{fn1}}
\fntext[fn1]{Tianwei Zhang, Dong Wei, and Mengmeng Zhu contributed equally.}

\author[1]{Shi \snm{Gu}\corref{cor1}}
\cortext[cor1]{Corresponding author}
\ead{gus@uestc.edu.cn}
\author[2]{Yefeng \snm{Zheng}}

\address[1]{School of Computer Science and Engineering, University of Electronic Science and Technology of China, Chengdu 611731, China}
\address[2]{Jarvis Research Center, Tencent YouTu Lab, Shenzhen 518057, China}

\received{XX}
\finalform{XX}
\accepted{XX}
\availableonline{XX}

\begin{abstract}
Self-supervised learning has emerged as a powerful tool for pretraining deep networks on unlabeled data, prior to transfer learning of target tasks with limited annotation. The relevance between the pretraining pretext and target tasks is crucial to the success of transfer learning. Various pretext tasks have been proposed to utilize properties of medical image data (e.g., three dimensionality), which are more relevant to medical image analysis than generic ones for natural images. However, previous work rarely paid attention to data with anatomy-oriented imaging planes, e.g., standard cardiac magnetic resonance imaging views. As these imaging planes are defined according to the anatomy of the imaged organ, pretext tasks effectively exploiting this information can pretrain the networks to gain knowledge on the organ of interest. In this work, we propose two 
complementary pretext tasks for this group of medical image data based on the spatial relationship of the imaging planes. The first is to learn the relative orientation between the imaging planes and implemented as regressing their intersecting lines. The second exploits parallel imaging planes to regress their relative slice locations within a stack. Both pretext tasks are conceptually straightforward and easy to implement, and can be combined in multitask learning for better representation learning. Thorough experiments on two anatomical structures (heart and knee) and representative target tasks (semantic segmentation and classification) demonstrate that the proposed pretext tasks are effective in pretraining deep networks for remarkably boosted performance on the target tasks, and superior to other recent approaches.
\end{abstract}

\begin{keyword}
\KWD Anatomy-oriented imaging plane\sep transfer learning\sep self-supervised pretraining
\end{keyword}

\end{frontmatter}

\section{Introduction}\label{sec:introduction}

Medical image analysis (MedIA) has important clinical applications, including diagnosis \citep{silveira2009comparison}, quantitative analysis \citep{wei2013comprehensive}, prognosis \citep{gonzalez2018disease}, therapy planning \citep{jackson2018deep}, and risk assessment \citep{klifa2010magnetic}.
Since manual analysis of medical image data in big amounts can be labor-intensive, time-consuming, and subjective, computer-aided automated methods are of great value.
Benefiting from the progress of deep learning techniques, especially the deep neural networks (DNNs), automated MedIA methods have advanced remarkably in recent years \citep{litjens2017survey}.
However, to achieve satisfactory performance, DNNs often need large amounts of labeled data for effective training, which can be difficult and costly to obtain in practice.

Transfer learning is an effective technique when the data is insufficient for training DNNs from scratch \citep{tan2018survey}, where the model is first pretrained on tasks with sufficient data and then fine-tuned on the target task with limited data and annotations.
For transfer learning on natural images, pretrained models on the ImageNet \citep{russakovsky2015imagenet} are available for a variety of popular DNN structures and are routinely used nowadays.
However, as the distance between the pretraining and target tasks plays a crucial role in transfer learning \citep{zhang2017self}, these models may become less effective when transferred to MedIA tasks due to the large gap between the two types of images.
Therefore, researchers are confronted with a dilemma.
On one hand, for effective transfer learning on medical images, models pretrained with medical images of highly relevant tasks are preferred.
On the other hand, it is difficult to obtain large quantities of annotations for medical images for the pretraining.
Fortunately, an emerging subfield of deep learning known as self-supervised learning (SSL) suggests a way out.

In SSL, the training tasks and supervision signals are defined by inherent properties of the data without any manual annotation \citep{jing2020self}.
Therefore, SSL can pretrain the models on a big amount of unlabeled data with proper \textit{pretext} tasks.
To exploit unique properties of medical image data, various pretext tasks have been proposed.
\citet{SpinalMRIs2017} proposed to train a Siamese network based on patient identity.
More recently, Models Genesis \citep{zhou2019models} and Rubik's cube series \citep{rubik2020self2} presented generic pretext tasks for medical image data, especially volumetric images.
However, previous works rarely paid attention to medical image data with anatomy-oriented imaging planes, {\color{black}e.g., magnetic resonance imaging (MRI) of various organs and body parts such as heart, knee, and shoulder,\footnote{\color{black}Readers are referred to https://mrimaster.com/ for more details as well as more anatomical structures adopting anatomy-oriented imaging planes.} which constitute a large portion of medical image data besides full 3D volumetric (e.g., CT) and 2D (e.g., X-ray) images.}
As such imaging planes are defined with respect to the anatomy of the imaged organ, pretext tasks that can effectively utilize this information are expected to be more relevant to potential target tasks (also known as downstream tasks) on the organ of interest than the generic ones.

In this work, we propose two pretext tasks for medical image data with anatomy-oriented imaging planes based on the spatial alignment relationship among multiple imaging planes.
The first is to learn the relative orientation between the imaging planes.
In clinical imaging, it is common to define anatomy-oriented view planes for obtaining comparable biometric measurements across populations.
These are called standard view planes for a specific organ.
For example, the neuro-imaging community defines the mid-sagittal plane in evaluation of pathological brains by estimating the departures from bilateral symmetry in the cerebrum \citep{stegmann2005mid}.
Similarly, standard views are used in cardiac magnetic resonance (CMR) imaging for quantification of cardiac volumetry, function, and blood flow \citep{kramer2020standardized}.
A key component of acquiring these standard views is the identification of specific anatomical landmarks to prescribe the imaging planes.
As a result, these imaging planes often intersect with each other at anatomically meaningful landmarks, and the intersecting lines can provide strong cues about the imaged organ.
Therefore, we propose to predict the intersecting lines between the imaging planes by regressing a distance-based heatmap.

Our second pretext task is complementary to the first, which exploits the spatial relationship among parallel imaging planes (in contrast with that between intersecting ones).
Specifically, we propose to regress the relative locations of the slices within a stack.
To solve this task, the network must gain an understanding of the within-slice content (focused on the imaged organ) and the cross-slice context,
thus is better prepared for potential downstream tasks on the specific organ.
Closely related to our work, \citet{zhang2017self} proposed pair-wise ordering of slices extracted from
volumetric scans, for the downstream task of fine-grained body part recognition.
However, their pretext task may encounter difficulty in handling objects with symmetrical structures, whereas we solve this problem with a small yet effective alteration, i.e., {\color{purple}centrosymmetric mapping}, for wider applicability.
Another difference is that we directly regress the relative slice locations with a single network, instead of ordering paired slices with a Siamese architecture.
In addition, we further investigate multi-task SSL combining both of the proposed pretext tasks, to fully exploit the two types of complementary spatial relationships among the imaging planes.

In summary, the prominent contribution of this work is the proposal of two complementary pretext tasks for 
self-supervised learning of medical image data with anatomy-oriented imaging planes.
{\color{purple}We hypothesize that these tasks are more relevant to potential downstream tasks than general transformation-and-recovery based pretext tasks \citep{zhou2019models, puzzle2016unsupervised} on such data, thus expected to lead to better transfer learning.}
Besides, they are conceptually straightforward and easy to implement.
For evaluation, we conduct thorough experiments for imaging-based analysis of two different anatomical structures (heart and knee) and two representative downstream tasks (semantic segmentation and classification).
We not only investigate the impact of the proposed pretext tasks on the downstream tasks but also study their learnability.
The results indicate that the proposed pretext tasks lead to better transfer learning than other recently proposed competitors, {\color{purple}empirically confirming our hypothesis}.

\section{Related work}\label{sec:literature}

\subsection{Transfer Learning}

Transfer learning~\citep{tan2018survey} is a powerful technique against shortage of training data, where the model parameters are first pretrained on a data-rich task before fine-tuned on a downstream task of limited data.
Its effectiveness has been demonstrated on diverse tasks \citep{yosinski2014transferable, transfer2014Vision, transfer2016Vision}, typically with a large labeled dataset like the ImageNet \citep{russakovsky2015imagenet} for pretraining.
For medical images, however, it is difficult to annotate data on such a large scale for supervised pretraining.
Although it is possible to transfer models pretrained on natural images to medical images \citep{transfer2016lung, transfe2016deep}, the huge domain gap between them may compromise the transfer efficacy \citep{Self-supervised2019}.
Therefore, pretraining methodologies that allow full exploitation of the large archive of medical image data in the absence of annotation are desirable.
Self-supervised learning (SSL)---an emerging deep learning paradigm that recently thrives---appears to be a promising solution.

\subsection{SSL for Natural Images}
In SSL, the supervision signal is defined by intrinsic properties of the data, eliminating the need for extrinsic annotation \citep{jing2020self}.
Therefore,  SSL is suitable for pretraining a model on large amounts of unlabeled data
\citep{chen2020big}.
Design of pretext tasks is crucial in SSL. An effective pretext task should make a good use of a certain data property, while being learnable yet non-trivial.
Various pretext tasks have been proposed and proved effective for natural images, including predicting relative positions of two image patches \citep{doersch2015unsupervised}, solving image jigsaw puzzles \citep{puzzle2016unsupervised}, colorizing grayscale images \citep{zhang2016colorful}, and predicting image rotations \citep{rotate2019self}.
{\color{black}Recently, we have witnessed a surge of SSL approaches based on contrastive/similarity learning of different views augmented from the same images \citep{chen2020simple,chaitanya2020contrastive,grill2020bootstrap}.
In addition, the masked autoencoder (MA) was proven successful as a scalable self-supervised vision learner, where the pretext task was to reconstruct the original image given its partial observation \citep{he2022masked}.
Despite their success, these pretext tasks were primarily designed for general-purpose transfer learning of especially natural images and did not consider the unique characteristics of anatomy-oriented medical image data.}

\subsection{SSL for Medical Images}
Among the differences between natural images and many medical images, one of the most notable is the {\color{purple}3D} spatial property of the latter.
Various pretext tasks have been proposed to exploit 3D information of volumetric medical images.
A series of pretext tasks of solving a Rubik's cube was constructed from volumetric input in \citep{rubik2020self2,rubik2020self3} for generic 3D medical image analysis.
Models Genesis \citep{zhou2019models} is another example of pretext tasks designed for generic 3D medical image analysis, where the input underwent a series of random transformations 
and the task was to restore the original input.
In contrast,  \citet{brain2018self} proposed to predict the geodesic distance along the brain surface between two patches randomly sampled from the cortex of the same brain using a Siamese network, {\color{purple}for improving cytoarchitectonic segmentation of human brain areas,}
which is an example of pretext tasks exclusively tailored for target downstream tasks.

However, none of the above-described pretext tasks utilized the spatial relationship of a set of intersecting slices with anatomy-oriented imaging planes, instead of a full 3D volume.
Since the intersecting imaging planes are prescribed concerning the structural landmarks of the imaged organ, pretext tasks based on their spatial relationship are expected to teach the networks about the anatomy of the organ.
{\color{black}In a pioneering work---in fact, the only one that we are aware of---along this line, \citet{bai2019self} proposed to define a segmentation pretext task based on the relative orientation between intersecting CMR planes and achieved encouraging results on the downstream task of accurate ventricle segmentation.
Concretely, they defined square boxes along the intersecting lines as the segmentation targets.
We share the same motivation of utilizing the spatial relationship between anatomy-oriented imaging planes for SSL but 
%
build upon the recent progress in DNN-based keypoint detection to regress a heatmap defined by the distance to the intersecting line, which can be considered a continuous and softened version of \citet{bai2019self}'s pretext task.}

\subsection{Keypoint Detection}
In keypoint detection \citep{zhou2018starmap}, the input image is fed to a fully convolutional network \citep{long2015fully} to generate a multi-channel heatmap, where each channel stands for a keypoint and the peaks indicate keypoint locations.
The network is trained in a fully supervised way by a
Gaussian heatmap defined with ground truth keypoint locations, where each keypoint defines the mean of a Gaussian kernel.
Besides the prevalent application to well-defined semantic keypoints, e.g., human joints \citep{newell2016stacked,NIPS2017_8edd7215}, there is a recent trend of general implicit keypoint detection applied to the task of object detection \citep{zhou2018starmap, law2018cornernet, zhou2019bottom}.
The keypoint-based objection detection eliminated two drawbacks of the anchor box based methods \citep{law2018cornernet}: (i) the imbalance between positive and negative anchor boxes and the resulting slow convergence, and (ii) the complicated hyperparameters and design choices of the anchor boxes, including number, size, and aspect ratio.
{\color{black}In this work, we extend the concept of DNN-based keypoint detection for line detection, i.e., the intersecting lines between two imaging planes.}

\subsection{Multi-Task SSL}
Joint learning of multiple related tasks has proven effective in learning more robust feature representations for better generalization, thus leading to improved performance \citep{MTL2017overview}.
Therefore, it is no surprise that multi-task learning (MTL) has also been actively explored in SSL.
\citet{MTL2017multi} explored multi-task SSL for natural images with four pretext tasks, and showed that even a naive multi-head architecture could achieve consistent improvement in performance.
As to medical image analysis, \citet{Multi2020self} proposed ColorMe, a framework combining spatial context and color distribution of scopy images for SSL.
However, ColorMe was proposed for color images and inapplicable to grayscale images such as magnetic resonance imaging (MRI).
In this work, we also explore multi-task SSL where the pretext task of regressing relative slice orientations is combined with a second pretext task of regressing relative slice locations.

\begin{figure*}[t]
\centering
\includegraphics[width=.8\textwidth,trim=0 0 0 0,clip]{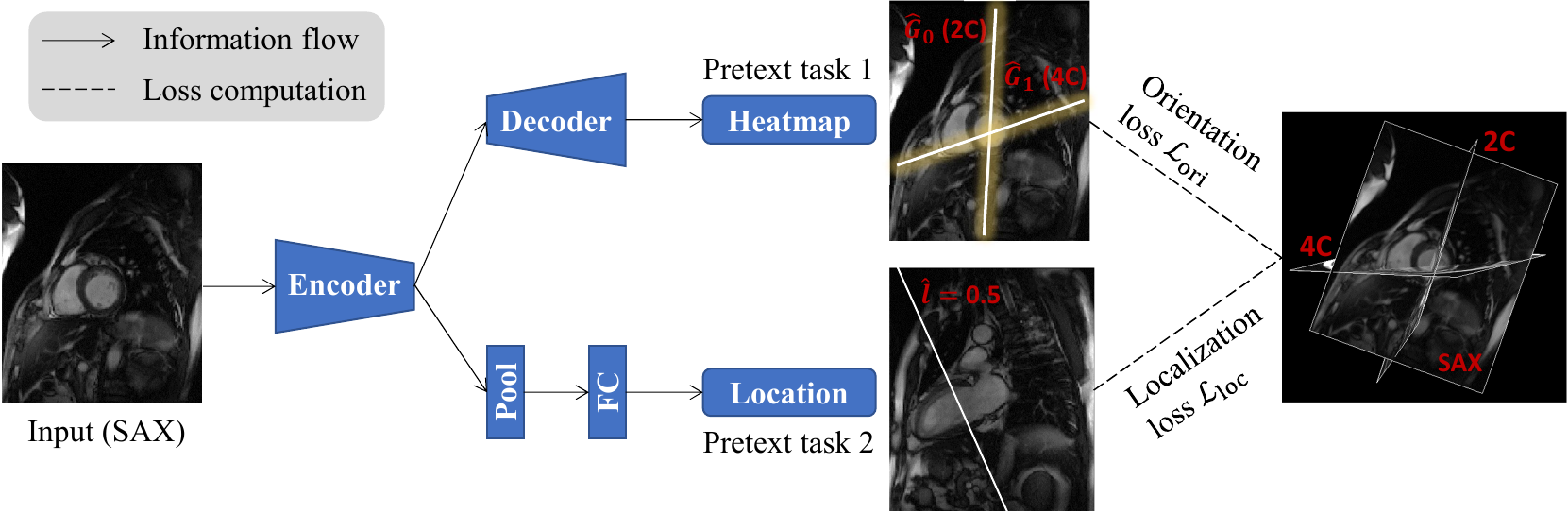}
\caption{\color{black}Network architecture for SSL of the proposed pretext tasks (FC: fully connected layer).
Top: relative orientation regression; bottom: relative location regression.}\label{fig:network}
\end{figure*}

\begin{figure}[t]
  \centering
  \includegraphics[width=\linewidth,trim=0 7 0 0,clip]{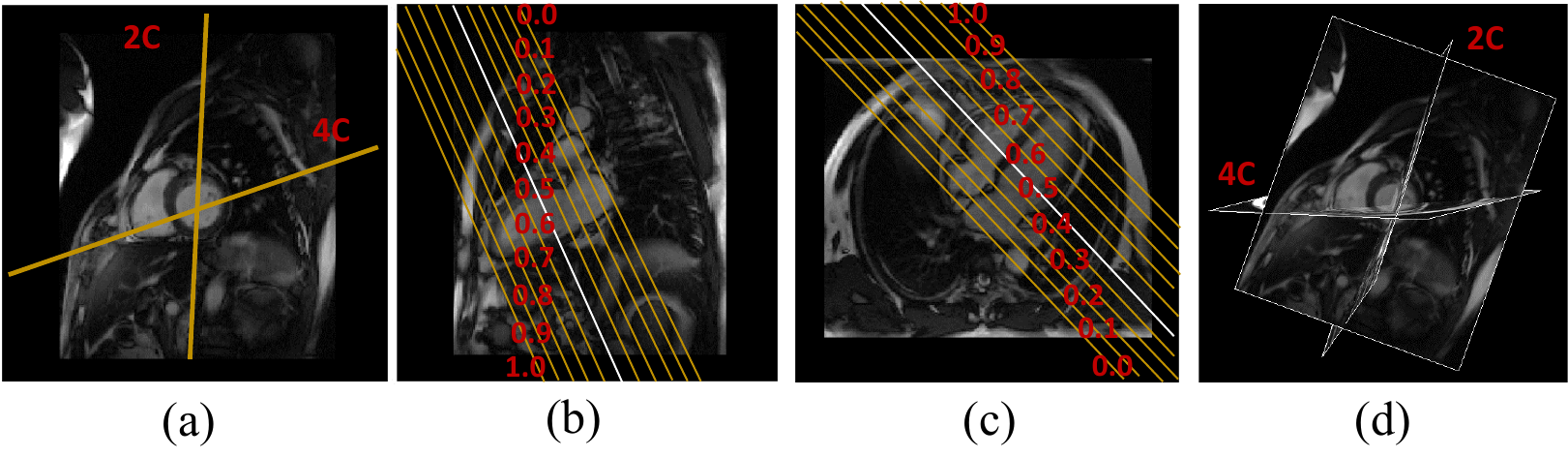}
  \caption{Standard CMR views.
  (a) A mid-ventricular SAX view: the straight lines are the intersecting lines with the 2C and 4C views in (b) and (c), respectively.
  (b)--(c) Standard 2C and 4C views: the parallel lines indicate intersecting lines with the stack of SAX views (with normalized relative locations marked), in which the white line indicates the SAX view in (a).
  (d) 3D visualization of the images in (a)--(c).}\label{fig:cmr_anatomy}
\end{figure}

\section{Preliminaries}\label{sec:preliminary}

As introduced earlier, anatomy-oriented imaging planes are commonly used in clinical practice for obtaining comparable biometric measurements across populations, which are referred to as the standard views for a specific organ.
Take CMR for example, which is the standard for quantification of cardiac volumetry, function, and blood flow \citep{gerche2013cardiac}.
Cardiac pathologies are often best evaluated along the principal axes of the heart in the long-axis (LAX) and short-axis (SAX) views, rather than the axial, coronal, or sagittal plane defined with respect to the body axes.
The most commonly used standard CMR views include a stack of SAX views (Fig. \ref{fig:cmr_anatomy}(a)), a two-chamber (2C; Fig. \ref{fig:cmr_anatomy}(b)) LAX view, and a four-chamber (4C; Fig. \ref{fig:cmr_anatomy}(c)) LAX view.
The SAX views are perpendicular to the long axis of the left ventricle (LV), whereas the LAX views are along the long axis. These views provide complementary information for a comprehensive evaluation of the heart.
Although different protocols are used for prescribing the standard SAX and LAX view planes by different vendors and institutions, the general consensus is that both the SAX and LAX views should be tailored to the unique individual anatomy \citep{kramer2020standardized}.
For example, the 4C plane should pass through the center of the LV and the right ventricle (RV) apex in the SAX views, whereas the 2C plane should bisect the LV while in parallel to the ventricular septum  (see the intersecting lines in Fig. \ref{fig:cmr_anatomy}(a)).
In addition, the stack of SAX views should cover the LV from the base through the apex with an even spacing (Figs. \ref{fig:cmr_anatomy}(b) and (c)).

The spatial information of a modern medical image is fully recorded when it is saved properly, e.g., in the Neuroimaging Informatics Technology Initiative\footnote{https://nifti.nimh.nih.gov/}
(NIfTI) or Digital Imaging and Communications in Medicine\footnote{https://www.dicomstandard.org/} (DICOM) format.
Particularly, DICOM is the international standard for medical images and related information, and is used by almost all radiographs nowadays.
It defines the formats for medical images that can be exchanged with the data and quality necessary for clinical use.
The DICOM header contains two attributes that record the location and orientation of a medical image:
(i) Image Position Patient (IPP): the $x$, $y$, and $z$ coordinates of the upper left corner (center of the first voxel transmitted) of the image with respect to the reference coordinate system (RCS), a patient-based coordinate system; and
(ii) Image Orientation Patient (IOP): the direction cosines of the first row and the first column of the image with respect to the RCS.
Using IPP and IOP, the spatial relationship between any two radiographs (of the same exam) can be readily computed.

\section{Methods}\label{sec:method}

In this section, we propose two novel pretext tasks---regressing relative orientations and relative locations---for medical images with anatomy-oriented view planes (Fig. \ref{fig:network}).
Both tasks are based on the spatial information self-contained in the radiographs, thus needing no manual annotation.
Meanwhile, both tasks require an understanding of the image contents to accomplish.
Therefore, learning them can prepare the networks for relevant downstream target tasks, e.g., semantic segmentation and classification as demonstrated in this work, leading to better transfer learning.
In addition, the two tasks can be combined for MTL, further boosting the SSL.
For consistency with the previous section, we continue to use CMR as the example for a description of the proposed pretext tasks.

\subsection{Relative Orientation Regression}

The first pretext task is to predict the relative orientation of an imaging plane within another intersecting one.
As illustrated in Fig. \ref{fig:cmr_anatomy}(a), the 2C LAX view bisects the LV while in parallel to the ventricular septum in SAX views, and the 4C LAX view bisects the LV while passing through the RV apex.
Given a SAX image as input, we propose to train the networks to predict its intersecting lines with the 2C and 4C LAX views, which can be readily computed using their IPPs and IOPs.
{\color{black}The intuition is that, to correctly predict the intersecting lines, the networks have to gain an understanding of the cardiac structures based on which the anatomy-oriented view planes are prescribed, such as the LV, RV, ventricular septum, and the myocardium.}
Once well trained by the pretext task, the networks are expected to be better prepared (pretrained) for potential downstream tasks, e.g., multi-structural segmentation.

To define the regression ground truth for each input SAX image, we first compute its intersection lines with both the 2C and 4C LAX view planes, obtaining two straight lines across the SAX image.
Then, a heatmap is constructed from each of these two lines (see Fig. \ref{fig:heatmap_CMR} for examples), based on the distance to the line and a Gaussian kernel:
\begin{linenomath*}
\begin{equation}\label{eq:heatmap}
    G(x,y)=\exp\big[-{(Ax+By+C)^2}/{\big(2\sigma^2(A^2+B^2)\big)}\big],
\end{equation}
\end{linenomath*}
where $(x,y)$ denotes the coordinate of a pixel, $Ax+By+C=0$ is the equation of the intersection line in the coordinate system of the SAX image, {\color{black}$A$, $B$, and $C$ are real-number coefficients in the standard form of the equation of a line,}
and $\sigma$ is the standard deviation of the Gaussian kernel.
{\color{purple}Through preliminary experiments, we find that the transfer learning performance on our downstream tasks is not sensitive to the exact value of $\sigma$, and simply fix it to 6 pixels in this work.}
Similar to the Gaussian-based heatmaps commonly used in the keypoint detection literature \citep{Pfister_2015_ICCV}, using the ``softened'' ground truth as the training target imposes less penalty when the prediction gets closer to the exact location, thus encouraging the prediction to gradually approach the desired status.
A fully convolutional network \citep{long2015fully} with the standard encoder-decoder architecture (e.g., the commonly used U-Net \citep{ronneberger2015u}) can be employed to regress the two heatmaps (Fig. \ref{fig:network} top).
The output has two channels, one for each heatmap. Similar to \citep{Pfister_2015_ICCV}, an L2 loss is employed to train the network:
\begin{linenomath*}
\begin{equation}\label{eq:L_heat}		
		\resizebox{0.95\hsize}{!}{$\begin{aligned}
		\mathcal{L}_\mathrm{ori} = \frac{1}{NK|\Omega|}
        {\sum}_{i=1}^N{\sum}_{k=1}^K{\sum}_{(x,y)\in\Omega}
        \big\|G_{i,k}(x,y) - \hat{G}_{i,k}(x,y)\big\|^2,
		\end{aligned}$}
\end{equation}
\end{linenomath*}
where $N$ is the total number of slices, $K$ is the total number of heatmaps to predict (also the number of output channels), $(x,y)$ iterates over all the pixels in the input image domain $\Omega$, and $\hat{G}$ is the predicted heatmap.

\subsection{Relative Location Regression}

\begin{figure}[t]
 \centering
 \includegraphics[width=.99\columnwidth,trim=0 0 0 0,clip]{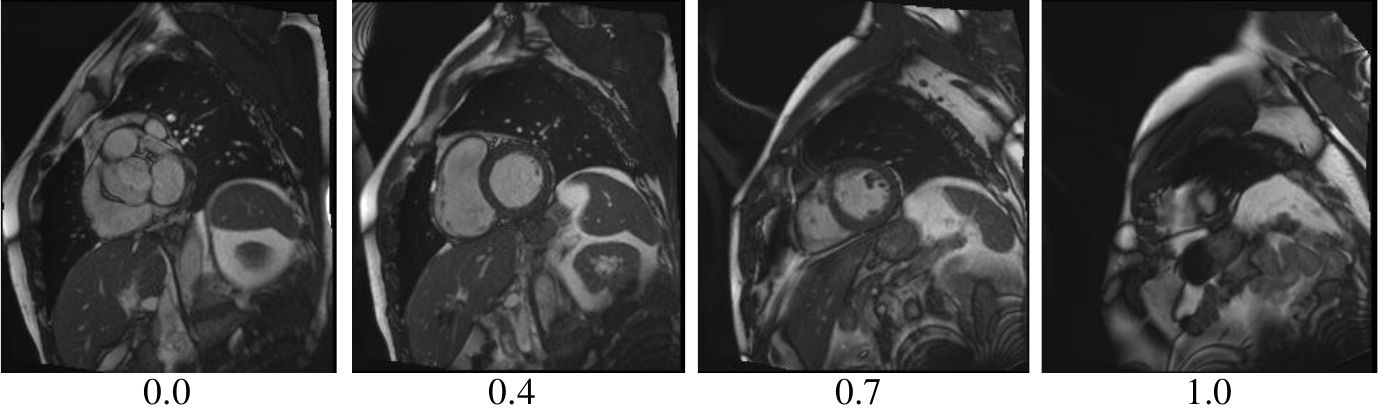}
 \caption{Left to right: four SAX CMR slices from the base (relative location$=$0.0) to the apex (relative location$=$1.0) of the LV.}\label{fig:cmr_sid}
\end{figure}

For medical imaging employing anatomy-oriented view planes, it is common that a stack of parallel slices
is prescribed to fully cover the structure of interest.
For example, in CMR, a stack of SAX views is prescribed to cover the LV from the base to the apex (Figs. \ref{fig:cmr_anatomy}(b) and (c)).
Naturally, the consecutive views reflect the gradual anatomical changes in sequence (Fig. \ref{fig:cmr_sid}), and a trained observer can speculate about the relative location of a specific slice by its appearance, with respect to the entire structure of interest.
Driven by this observation, the second pretext task we propose is to predict the relative location of an input slice within the parallel stack, which is complementary to the first pretext task.

Practically, we consider two situations when defining the relative location.
The first situation is just like the CMR illustrated above, in which we define the relative location (denoted by $l$) of a specific slice to be its normalized location within the stack:
\begin{linenomath*}
\begin{equation}\label{eq:relative_loc}
  l_{s_i} = d(s_i, s_1)/d(s_N, s_1),
\end{equation}
\end{linenomath*}
where $s_i$ is the $i$\textsuperscript{th} slice, $N$ is the total number of slices of the stack, and $d(\cdot, \cdot)$ is the Euclidean distance between two parallel slices.
{\color{purple}Hence, the relative locations of the entire stack of slices is normalized to span the range of $[0, 1]$ for different subjects.}
In the second situation, the slices may be symmetrically similar about a ``mirror'' point.
For example, in sagittal knee MRI, the slices on both sides of the middle slice can be difficult to differentiate.
In this case, we use the sine function to map the raw distance ratio by:
\begin{linenomath*}
\begin{equation}\label{eq:sym_loc}
  l_{s_i} = \sin \big[\pi \cdot {d(s_i, s_1)} / {d(s_N, s_1)} \big].
\end{equation}
\end{linenomath*}
In this way, $l_{s_i}$ becomes symmetric about the central slice while still spanning the full range of $[0, 1]$, with symmetric slices on different sides of the mirror point mapped to similar values,
%
and the pretext task is made more generically applicable.
Again, the relative slice locations are defined by mining the spatial attributes recorded in the DICOM header. Therefore, no manual annotation is needed.

Given an input slice $s_i$, we let the network to regress its relative location and define the loss function as:
\begin{linenomath*}
\begin{equation}\label{eq:L_sid}
    \mathcal{L}_\mathrm{loc}=\frac{1}{N}{\sum}_{i=1}^{N}\big\|l_{s_i} - \hat{l}_{s_i}\big\|^2,
\end{equation}
\end{linenomath*}
where $\hat{l}_{s_i}$ is the network's prediction.
Implementation-wise, we pass the feature map obtained at the end of the encoder through a pooling layer to convert it to a feature vector, followed by a fully connected layer to regress the relative location (Fig. \ref{fig:network} bottom).
We expect that training the network with this pretext task should also prepare it for relevant downstream tasks, since visually speculating about the relative slice locations requires knowledge about the organ of interest and its surroundings.

\subsection{Multitask SSL}
We further explore integration of the two complementary pretext tasks for multitask SSL, considering that both tasks require certain understanding of the organ of interest.
With better pretraining via MTL, the network is expected to yield better performance when transferred and fine-tuned on potential downstream tasks.
The loss function for the MTL is straightforward, i.e., the summation of the losses defined individually for the two pretext tasks:
\begin{linenomath*}
\begin{equation}\label{eq:L_mtl}
   \begin{aligned}
       \mathcal{L}_\mathrm{MTL} = \mathcal{L}_\mathrm{ori} + \mathcal{L}_\mathrm{loc}.
   \end{aligned}
\end{equation}
\end{linenomath*}
{\color{purple}In this work, an equal weight of 1 is used for $\mathcal{L}_\mathrm{ori}$ and $\mathcal{L}_\mathrm{loc}$ as we notice that they are generally in the same magnitude.
It is also worth mentioning that the pretext tasks are not dependent on any specific scanner manufacturer or model, as the relative orientation and location as supervision signal are computed only between slices scanned in the same study by the same machine.}

\subsection{Fine-Tuning on Target Tasks}
After the networks are well trained with the proposed pretext tasks, they can be fine-tuned with limited annotations for better performance on target downstream tasks.
In this work, we evaluate the efficacy of the proposed pretext tasks for transfer learning on two fundamental downstream tasks in medical image analysis: semantic segmentation and classification (i.e., imaging-based diagnosis).
For the former, both the pretrained encoder and decoder are used, with the regression head replaced by a multiclass segmentor consisting of a 1$\times$1 convolution layer followed by softmax.
We use the cross-entropy loss for training.
For the latter, only the pretrained encoder is needed.
We reinitialize the fully connected layer and affix a sigmoid function to convert the output to classification probabilities.
A weighted binary cross-entropy loss is employed
to account for imbalanced class sizes, in which the loss for a sample is scaled inversely proportional to the prevalence of its class in the dataset.
{\color{black}The encoder and decoder parameters are adjusted during fine-tuning instead of fixed.}

\section{Experiments}\label{sec:exp}

\subsection{Materials}
To evaluate the efficacy of the proposed self-supervising pretext tasks on transfer learning of medical image analysis of different body parts, we use cardiac and knee MRI for thorough experiments, which are representative of medical image data with anatomy-oriented imaging planes.

\subsubsection{CMR}

For self-supervised pretraining, the Data Science Bowl Cardiac Challenge\footnote{https://www.kaggle.com/c/second-annual-data-science-bowl/overview} (DSBCC) dataset is used.
The dataset comprises cine images of more than 1,000 subjects in DICOM format, with a diverse representation of individual variations, {\color{purple}including subjects from young to old, images from numerous health centers, and hearts of normal to abnormal cardiac function}.
{\color{black}The images were acquired with two scanners (Siemens Area 1.5 T and Siemens Skyra 3.0 T).}
The official training, validation, and test sets include 500, 200, and 440 CMR exams, respectively.
Our pretext tasks here are to regress the relative orientations of the 2C and 4C views in the SAX views, and regress the relative locations of the SAX views.
Accordingly, the inclusion criterion is that the exam should include all of the SAX, 2C, and 4C views.
Thus, 451, 176, and 381 exams are included in this work from the official training, validation, and test sets, respectively.
In addition, we combine the official training and validation data for training, and use the official test data for validation of whether the pretext tasks can be successfully learned by the networks.
%
As no segmentation annotation is provided, the DSBCC dataset is only used for self-supervised pretraining.

For the downstream target tasks---multi-structural segmentation {\color{black}and abnormality diagnosis} in SAX CMR images, we use the Automated Cardiac Diagnosis Challenge (ACDC) dataset \citep{bernard2018deep}.
The dataset consists of 100 CMR exams acquired with two scanners (Siemens Area 1.5 T and Siemens Trio Tim 3.0 T)
at the University Hospital of Dijon, including {\color{black}five evenly distributed subgroups: normal, previous myocardial infarction, dilated cardiomyopathy, hypertrophic cardiomyopathy, and abnormal right ventricle.}
The cine images include a series of SAX views covering the LV from the base to the apex, without any LAX view though.
Images at the end-systolic (ES) and end-diastolic (ED) phases are provided with manual annotations of three cardiac structures (LV, RV, and myocardium).\footnote{\color{purple}The ACDC dataset does not record the spatial information, thus cannot be used for the proposed pretraining.}
As the official test set of the ACDC challenge is not publicly available, we divide the official training set into training, validation and test sets comprising 64, 16 and 20 subjects, respectively.

\begin{table}[t]
\centering
\caption{Acquisition protocols of the CMR and knee MRI datasets: DSBCC, ACDC~\citep{bernard2018deep}, fastMRI~\citep{zbontar2018fastmri}, and Stanford~\citep{MRnet}.}\label{tab:protocol}
\setlength{\tabcolsep}{.8mm}
\begin{adjustbox}{width=\linewidth}
\begin{tabular}{cccccc}
\hline
\multirow{2}{*}{Dataset} & \multicolumn{2}{c}{CMR}  &  & \multicolumn{2}{c}{Knee}    \\ \cline{2-3} \cline{5-6}
                         & DSBCC & ACDC &  & fastMRI & Stanford \\ \hline
Width (pixel)            & 166--704    & 154--512   &  & 182--1024    & 192--224     \\
Height (pixel)           & 166--704    & 154--428   &  & 192--1024    & 384--512     \\
Field strength (tesla)   & 1.5 or 3.0  & 1.5 or 3.0 &  & 1.5 or 3.0   & 1.5 or 3.0   \\
Pixel spacing (mm)       & 0.60--1.80  & 0.70--1.92 &  & 0.146--1.237 & 0.293--0.313 \\
Slice thickness (mm)     & 4--11       & 5--10      &  & 2.0--5.0     & 2.5--3.5     \\
Slice gap (mm)     & 7.98, 8, 10 & 0 or 5     &  & 2.20--6.25   & 0, 0.5, 1    \\
No. frames per cycle     & 9--30       & 28--40     &  & -         & -         \\
No. slices per stack     & 8--18 & 6--18 &  & 15--62        & 24--42       \\ \hline
\end{tabular}
\end{adjustbox}
\end{table}

More details about the image acquisition protocols are presented in Table \ref{tab:protocol}.
We unify the in-plane resolutions of all SAX images in the two datasets via cubic interpolation to 1.260$\times$1.260 mm and 1.367$\times$1.367 mm, respectively, according to the modes of the image resolution distributions.
We then unify the size of the resampled images to 224$\times$224 pixels, either by central cropping
or zero padding.
Again, the image size is determined according to the mode of the image size distribution, after the resolution is unified.
Lastly, the z-score standardization is conducted on each image to make the pixel intensities have a zero mean and unit standard deviation.
{\color{black}We do not unify the slice thickness.}

\subsubsection{Knee MRI}

Knee MRI is the standard-of-care imaging modality for diagnosis of knee injuries \citep{naraghi2016imaging}.
{\color{black}In routine protocols, the axial and coronal planes should be perpendicular and parallel to the middle line of the femur and tibia in the sagittal views, respectively (see Fig. \ref{fig:heatmap_knee} for examples).}
Accordingly, our pretext tasks are to regress the orientations of the central coronal and axial slices within the sagittal images, and the relative locations of the stack of sagittal images.

For self-supervised pretraining, we use the data released as part of the fastMRI dataset by \citet{zbontar2018fastmri}.
This dataset includes DICOM data from 10,000 clinical knee MRI exams, representing a wide variety of scanners and pulse sequences.
Each exam typically contains several sequences;
as only the sagittal T2W with fat suppression is also included by the protocol of the dataset for the target task (described next), thus it is selected for our pretext tasks.
The official Batch 1 and Batch 2 are used for training and validation, respectively.
Only the exams that include sagittal T2W images accompanied by both axial and coronal views are included.
The fastMRI dataset has no label for training or evaluation of any target task of diagnosis.

On the Stanford dataset \citep{MRnet}, we target at the downstream task of detecting general abnormalities and specific diagnoses (anterior cruciate ligament (ACL) and meniscal tears) in knee MRI.
The examinations were performed at the Stanford University Medical Center
with GE scanners (GE Discovery, GE Healthcare, Waukesha, WI) and
a routine non-contrast knee MRI protocol, including the sagittal T2W with fat saturation sequence.
%
We combine the original training (1,130 exams) and tuning (120 exams) sets (the test set is not publicly available), and randomly divide them into our training, validation, and test sets of 800, 200, and 250 exams, respectively, while maintaining the ratios of disorders (80.6\% abnormal, including 21.0\% with ACL tears and 35.9\% with meniscal tears).\footnote{\color{purple}The Stanford dataset does not record the spatial information, thus cannot be used for the proposed pretraining.}

More details about the image acquisition protocols are presented in Table \ref{tab:protocol}.
Similar to the CMR data, the two knee datasets are unified via interpolation (to 0.5$\times$0.5 mm pixel spacing), cropping or zero-padding (to 256$\times$256 pixels), and z-score standardization.
{\color{black}Again, the slice thickness is not unified.}

\subsection{Evaluation Metrics}

\subsubsection{Pretext Tasks}
For the task of relative location regression, the mean squared error (MSE), mean absolute error (MAE), Pearson correlation coefficient (denoted by $r$), coefficient of determination (denoted by $R^2$), and explained variance \citep{demaris2002explained} are used to evaluate the learning outcome.
In addition, linear regression analysis between the predicted and ground truth values is performed.
For the task of regressing relative orientation, we resort to visual inspection of the predicted heatmaps in comparison with the ground truth.

\subsubsection{Target Tasks}
For multi-structural segmentation in CMR, the Dice coefficients and average symmetric surface distance (ASSD) are computed volume-wise for the LV, RV, and myocardium, in accordance with the ACDC challenge \citep{bernard2018deep}.
In addition, mean values averaged across the three structures are computed for straightforward comparison among methods.
For diagnosis of {\color{black}cardiac and} knee MRI, the area under the curve (AUC) of the receiver operating characteristic curve is employed,
which can better evaluate classification performance on imbalanced data.
Specifically, for knee MRI, we treat the diagnosis of each of the three abnormalities (general abnormalities, ACL tears, and meniscal tears) as a binary classification task following \citet{MRnet} and report the mean AUCs across the three tasks.

\subsection{Implementation}
We use the PyTorch framework \citep{paszke2017pytorch} and Adam \citep{kingma2014adam} optimizer for all experiments.
A single NVIDIA TITAN Xp GPU is used for training and inference.
{\color{purple}Our code will be available.}

For CMR, we employ the U-Net architecture \citep{ronneberger2015u} as backbone for consistency with \citep{bai2019self}, a closely related work which also experimented on the ACDC dataset and we should compare to.
Mini-batches of 20 and 10 images are used for the pretext and target tasks, respectively.
The initial learning rate is set to 0.001.
For the pretext tasks, it is halved every 100 epochs, and we train for 500 epochs in total.
%
For the target task of multi-structural segmentation, the learning rate is halved every 50 epochs, and we train for a total of 200 epochs with an {\color{purple}L1 regularization} of $5\times10^{-5}$.
Online data augmentation, including random rotation and scaling, is performed for both the pretext and target tasks.
{\color{black}For the target task of abnormality diagnosis, we adopt the solution of the competition champion \citep{khened2018densely} to construct a 100-tree random forest classifier using ten features: ejection fractions of left and right ventricles, volumes of the LV at ES and ED, volumes of the RV at ES and ED, masses of the myocardium at ES and ED, and patient height and width.
The first eight features were computed with the ground truth segmentation maps to train the classifier.
Then, for testing, these features were calculated from the automatic segmentation results of the first target task.
}

For diagnosis of knee MRI, we employ the work \citep{MRnet} which published the Stanford dataset as our baseline, and follow it to use the AlexNet \citep{krizhevsky2012imagenet} as backbone.
Considering the availability of the ImageNet pretraining \citep{russakovsky2015imagenet} for the AlexNet, we stack the T2W images three times as the input to make use of the ImageNet pretrained parameters, and further pretrain (fine-tune) the network with the SSL pretext tasks.
{\color{purple}Following \citet{MRnet},} slices of a T2W series are used as a mini-batch.
The total number of training epochs is set to 100 for both the pretext and target tasks.
For the pretext tasks, the learning rate is initially set to $1\times10^{-4}$, and decreases at the 50\textsuperscript{th} epoch by a factor of 0.1.
A 
{\color{purple}weight decay} of $5\times10^{-5}$ is used.
For the target task, the learning rate is initially set to $1\times10^{-5}$, and reduced when the validation loss fails to improve for five epochs in a row by a factor of 0.3, and the {\color{purple}weight decay}
is set to $1\times10^{-2}$.
{\color{black}Online data augmentation, including random rotation, shift, and horizontal flip,
is performed for both the pretext and target tasks.}
%

\begin{figure*}[t]
\centering
\includegraphics[width=.99\textwidth,trim=0 0 0 0,clip]{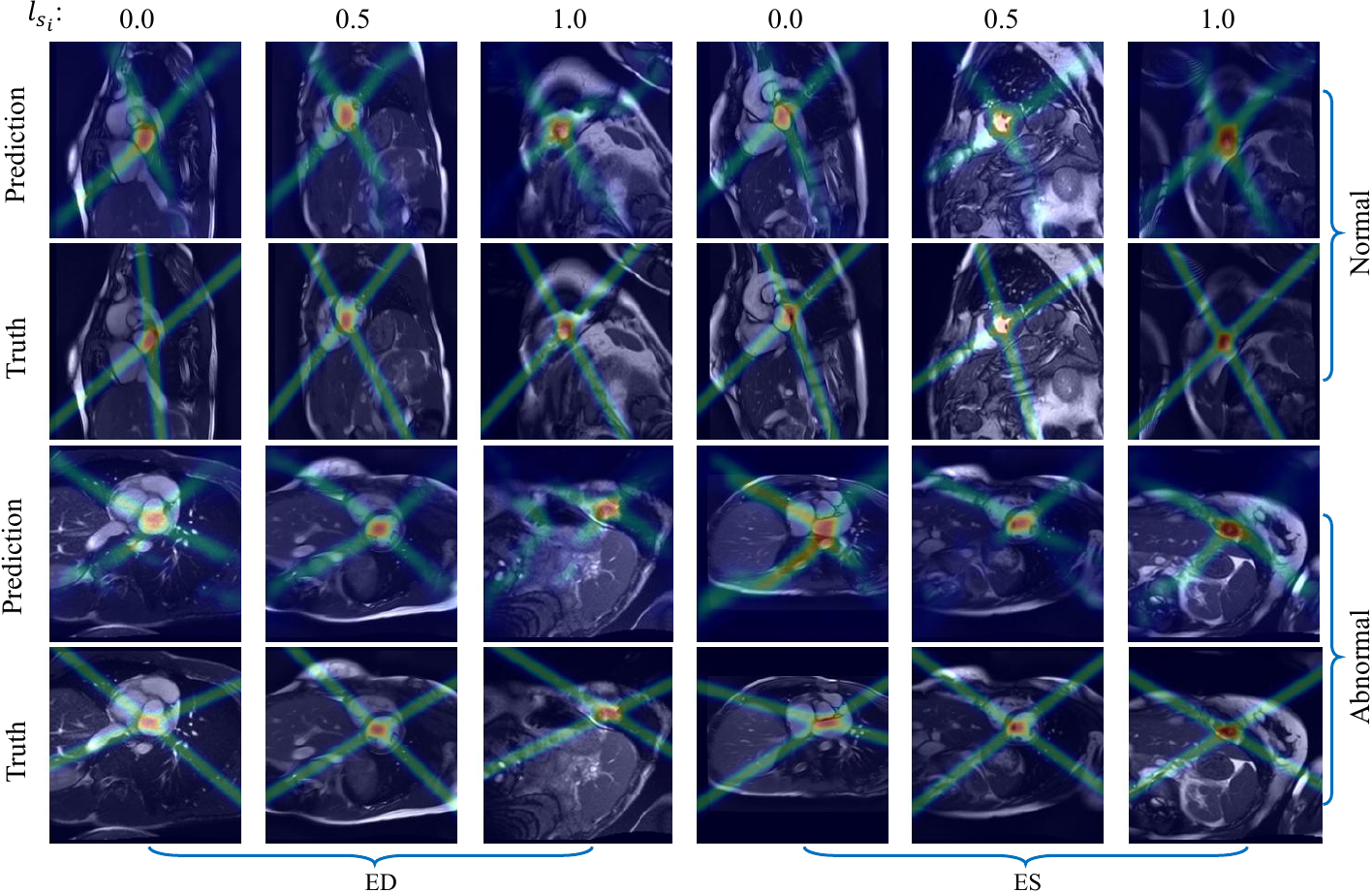}
\caption{\color{black}Visualization of the predicted heatmaps for the pretext task of relative orientation regression on the DSBCC CMR validation set, along with the ground truth.
ED: end-diastolic, ES: end-systolic.
Best viewed when digitally zoomed in.}\label{fig:heatmap_CMR}
\end{figure*}

\subsection{Experimental Settings}
The effectiveness of self-supervising pretext tasks is often evaluated by the transfer learning performance on downstream target tasks~\citep{zhou2019models,rubik2020self2},
and we follow this paradigm.
Specifically, we use different portions of the training data of the target tasks for training and compare the performance of (i) training from scratch and (ii) fine-tuning the parameters pretrained with the pretext tasks.
For the proposed pretext tasks, we study three pretraining settings: regressing relative orientations (Eqn. (\ref{eq:L_heat})), regressing relative locations (Eqn. (\ref{eq:L_sid})), and regressing both together (Eqn. (\ref{eq:L_mtl})).
We compare our pretext tasks with several established ones in the literature,
including the jigsaw puzzle \citep{puzzle2016unsupervised}, pairwise slice ordering \citep{zhang2017self}, Models Genesis \citep{zhou2019models},  anatomical position prediction (APP) \citep{bai2019self}, {\color{blue}SimCLR \citep{chen2020simple}, BYOL \citep{grill2020bootstrap}, and masked autoencoder \citep[MA;][]{he2022masked}.}
{\color{blue}For a fair comparison, we
adopt the same backbone networks, pretraining principles, and fine-tuning protocols for all the compared methods.
Precisely, for pretraining, we follow the recipes described in the original papers and pretrain the networks until convergence.
For fine-tuning, the protocols described in the previous section are applied to all methods.
}

{\color{blue}In addition, given that the pretrained weights on large-scale natural image datasets are widely available for the more recent SimCLR, BYOL, and MA approaches, we also evaluate the performance of transferring the pretrained weights of these methods with large-scale natural image pretraining for reference.
Specifically, we adopt the ImageNet \citep{russakovsky2015imagenet} pretrained ResNet-50 \citep{he2016deep} model for SimCLR and BYOL, and the ImageNet pretrained ViT-Base \citep{dosovitskiy2021an} model for MA.
For CMR segmentation, a U-Net-like segmentation network is built with the pretrained ResNet-50 and ViT-Base backbones;
whereas for knee MRI diagnosis, the pretrained backbones are directly used.
These ImageNet pretrained models will be marked with $\dagger$ signs when presenting experimental results later.
Again, the same protocols described in the previous section are employed for fine-tuning.
}

\begin{figure*}[t]
\centering
\includegraphics[width=.95\textwidth,trim=0 0 0 0,clip]{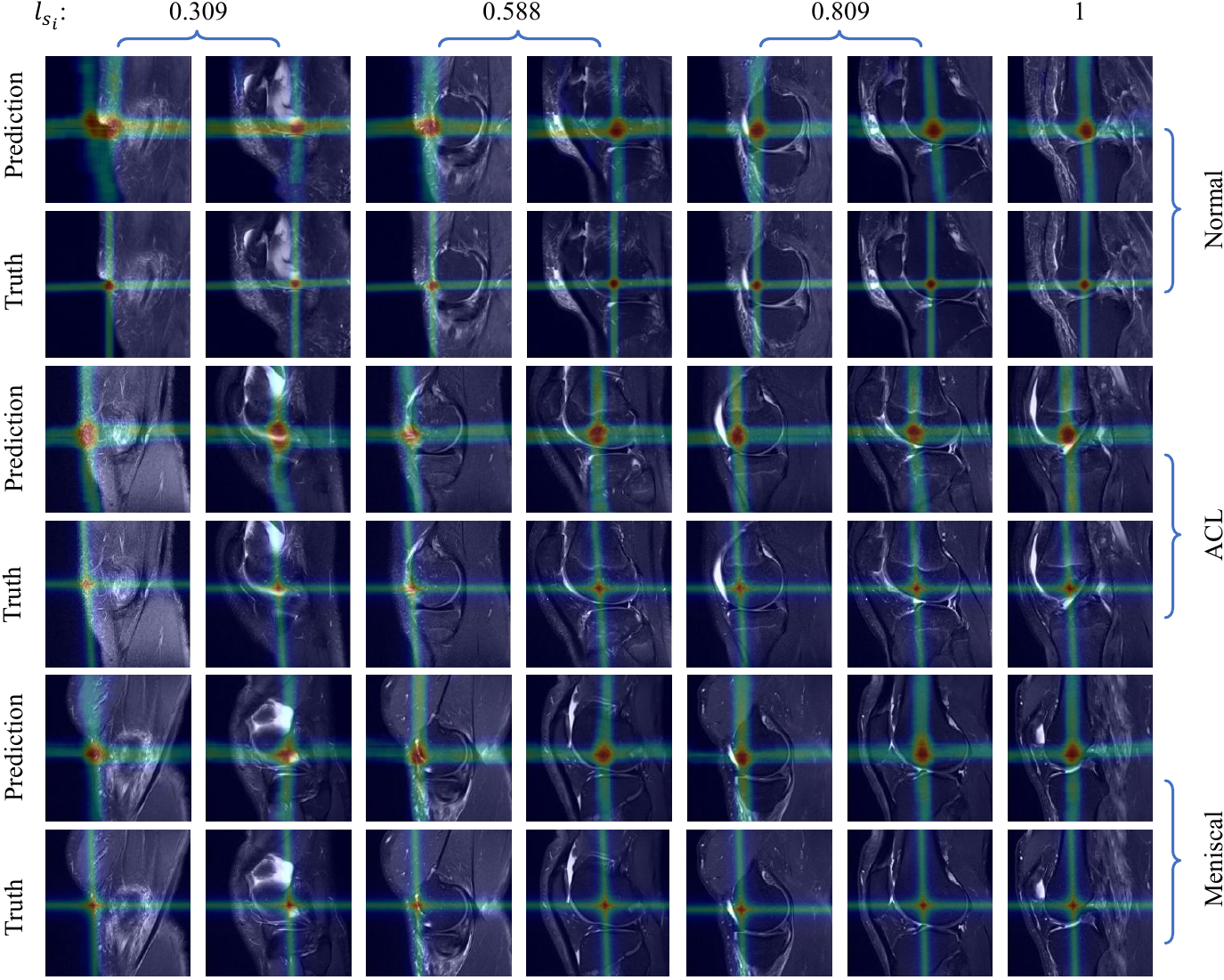}
\caption{\color{black}Visualization of the predicted heatmaps for the pretext task of relative orientation regression on the knee MRI validation data \citep{zbontar2018fastmri}, along with the ground truth.
Note that symmetric slices on different sides of the central slice are mapped to the same relative locations ($l_{s_i}$) by Eqn. (\ref{eq:sym_loc}).
ACL: anterior cruciate ligament.
Best viewed when digitally zoomed in.}\label{fig:heatmap_knee}
\end{figure*}

\subsection{Performance on Proposed Pretext Tasks}\label{sec:exp:perf_pre}

\subsubsection{Relative Orientation Regression}
We first visually inspect the predicted heatmaps for the intersecting lines between imaging view planes for both the cardiac and knee MRI datasets, in comparison with the ground truth.
Fig. \ref{fig:heatmap_CMR} displays SAX CMR slices of different cardiac phases, relative locations, and health status (normal and abnormal).
Fig. \ref{fig:heatmap_knee} displays sagittal knee MRI slices of different locations (on both sides of the knees) and health status.
As we can see, the models predict heatmaps consistently similar to the ground truth on both datasets, irrespective of the varying conditions.
The results suggest that the models have gained knowledge about the important anatomical landmarks that define these imaging planes via training with the proposed pretext task.

\begin{table}[t]
\centering
\caption{Validation results of the pretext task of regressing relative slice locations with single-task learning (Loc.) and multitask learning (MTL).
$\uparrow$: higher is better; $\downarrow$: lower is better.
MSE: mean squared error, MAE: mean absolute error, $r$: Pearson correlation coefficient, $R^2$: coefficient of determination, var.: variance.}
\label{tab:loc}
\setlength{\tabcolsep}{1.mm}
\begin{adjustbox}{width=\linewidth}
\begin{tabular}{ccccccc}
\hline\hline
Dataset  & Task & MSE$\downarrow$ & MAE$\downarrow$  &$r$$\uparrow$  & $R^2$$\uparrow$  & Explained var.$\uparrow$\\
\hline
DSBCC  & Loc. & 0.0187 & 0.1030 & 0.9187 & 0.8266 & 0.8304 \\
       & MTL & 0.0128 & 0.0826 & 0.9415 & 0.8822 & 0.8847 \\
\hline
fastMRI & Loc. &0.0146 &0.0914 &0.9254 &0.8555 &0.8558\\
        & MTL &0.0142 &0.0896 &0.9284 &0.8596 &0.8598\\
\hline\hline
\end{tabular}
\end{adjustbox}
\end{table}

\begin{figure}[t]
\centering
\includegraphics[height=4.1cm,trim=0 0 0 0,clip]{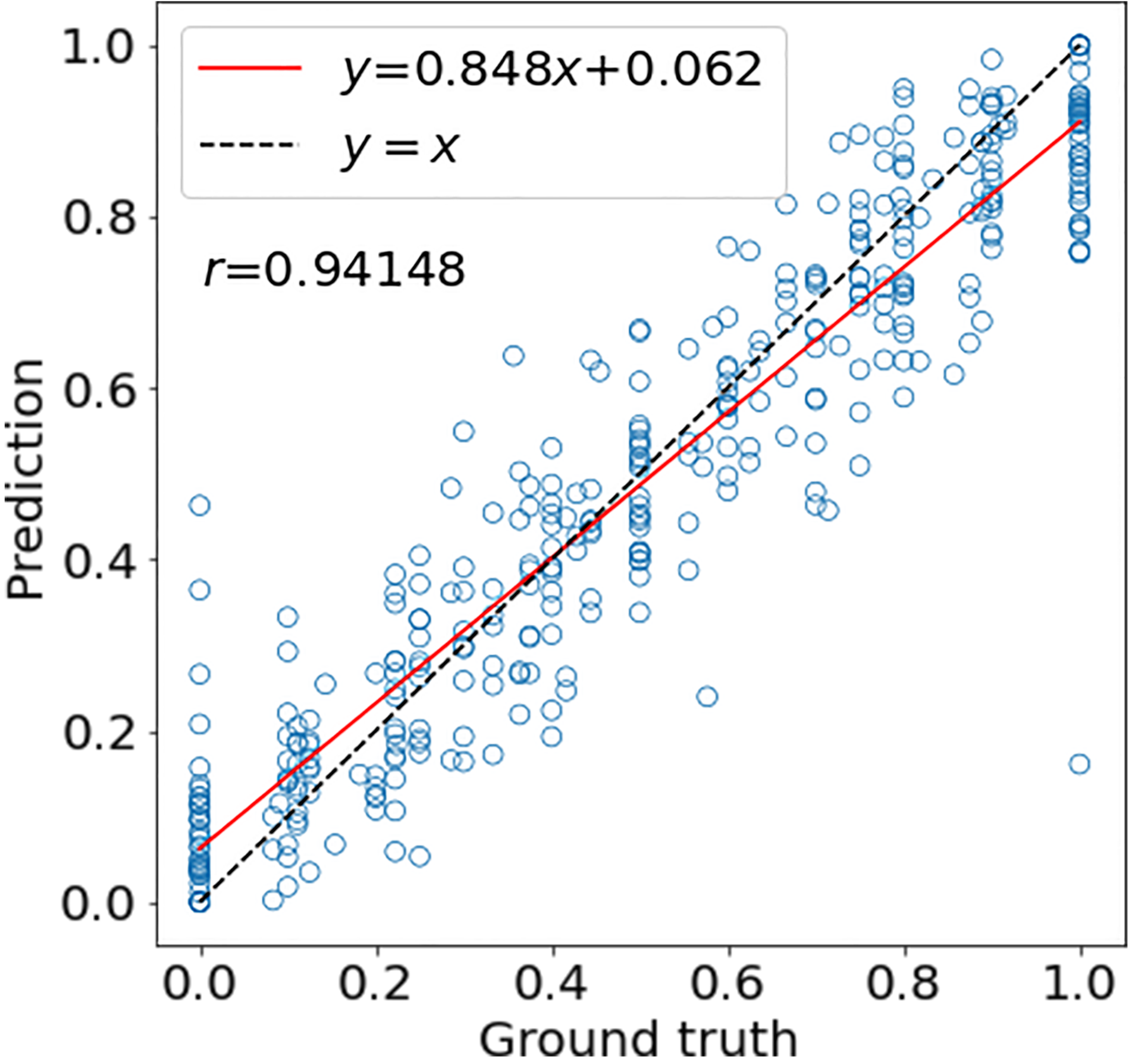}\hspace*{2mm}
\includegraphics[height=4.1cm,trim=60 0 0 0,clip]{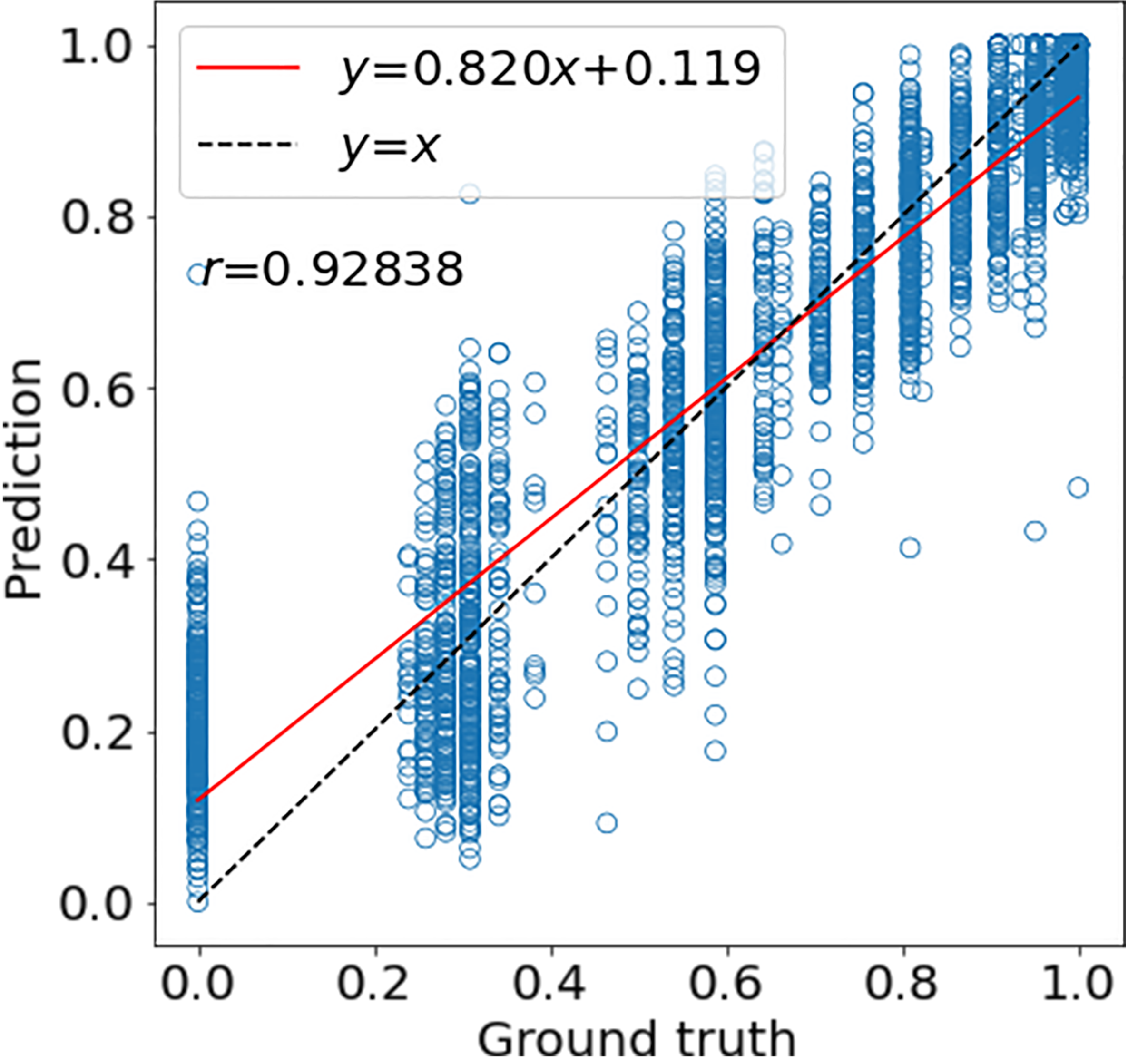}
\caption{Linear regression analyses between the predicted relative slice locations and ground truth.
Left: Cardiac MTL, and right: knee MTL.
}\label{fig:linear_reg}
\end{figure}


\begin{figure*}[t]
\centering
\includegraphics[width=\textwidth,trim=0 0 0 0,clip]{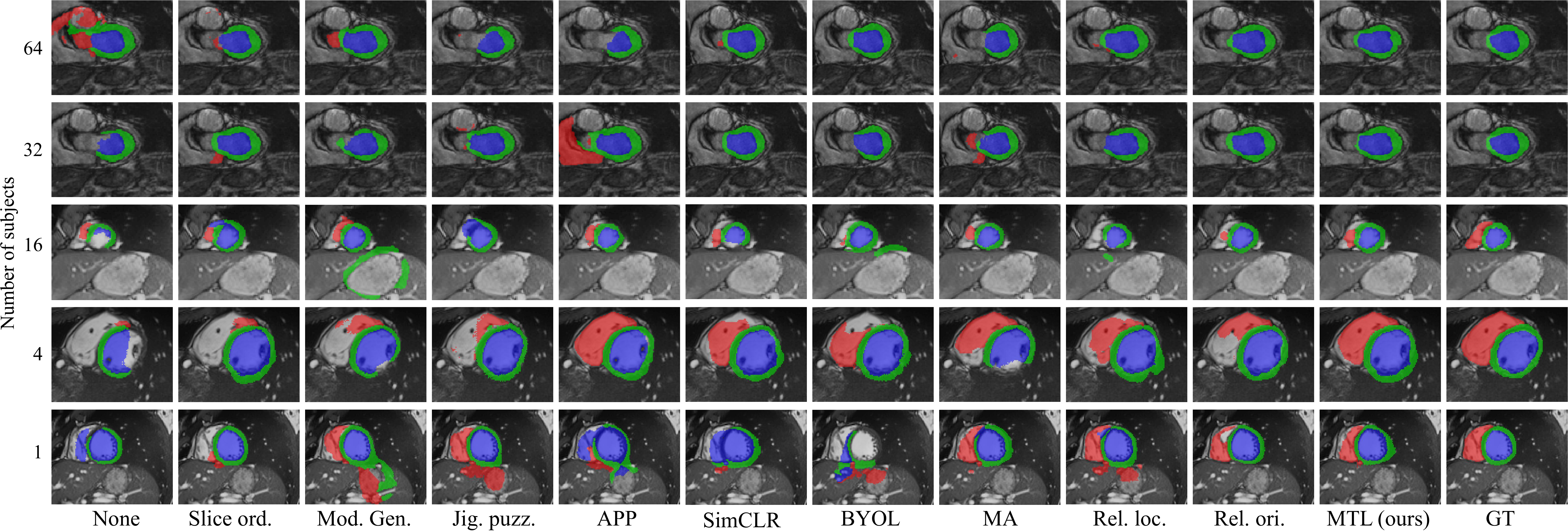}
\caption{\color{black}Visualization of the multi-structural segmentation results (red: RV, blue: LV, and green: myocardium) in SAX CMR images, corresponding to the results presented in Table \ref{tab:cmr}.
Note that to highlight the differences between the segmentation results with different pretext tasks, more challenging slices are used for visualization when more subjects are available for fine-tuning.
Rows 1 and 2: ES, and rows 3--5: ED.
}\label{fig:cmr_results}
\end{figure*}

\subsubsection{Relative Location Regression}
We then quantitatively validate the results of predicting relative slice locations in two settings: single-task and multitask learning (MTL).
The results are presented in Table \ref{tab:loc}.
As we can see, the task is performed well individually. All the evaluated metrics are fairly good on both datasets: the MSEs and MAEs are at the levels of 0.01 and 0.1, respectively, the Pearson coefficients are above 0.9, and both the coefficients of determination and explained variances are above 0.8.
These results suggest that the second pretext task can also be effectively learned by the network, thus embedding knowledge about the imaged structures of interest.
When further combining it with the other proposed pretext task, all the metrics improve on both datasets.
The linear regression analyses are plotted in Fig. \ref{fig:linear_reg}, which are in harmony with the results in Table \ref{tab:loc}.

\subsection{Transfer Learning Performance on Target Tasks}
\subsubsection{CMR Semantic Segmentation {\color{black}and Diagnosis}}
Table \ref{tab:cmr} presents the test performance
for the segmentation of the LV, RV, and myocardium on the ACDC dataset, as well as the cross-structure mean results.
From the table, we have the following observations.
First, the performances achieved by fine-tuning the networks pretrained with different pretext tasks are almost always better than those by training from scratch {\color{black}(except for the contrastive learning based approaches with extremely low data for fine-tuning, i.e., one subject)}, demonstrating the effectiveness of the self-supervising pretext tasks in transfer learning.
Second, the performance improvements are more obvious when fewer data are available for fine-tuning, emphasizing the prominent efficacy of self-supervised pretraining in the low-data regime.
Third, our multitask learning (MTL) integrating the relative orientation and location regression tasks achieves the greatest improvements in the majority of settings for both metrics and different anatomic structures.
When only four subjects are used for fine-tuning, our MTL achieves a reasonable performance of mean Dice at 0.814 and mean ASSD at 1.917 mm, which are comparable to training from scratch with 16 subjects (mean Dice at 0.815 and mean ASSD at 2.064 mm).
%
%
When 32 subjects are used, the MTL achieves considerably better performance than training from scratch with twice as much data.
Lastly, of the two proposed pretext tasks, relative orientation regression seems to be slightly more effective than relative location regression for the specific downstream task.

{\color{blue}Meanwhile, we note that the performances of directly transferring and fine-tuning ImageNet pretrained SimCLR$^\dagger$, BYOL$^\dagger$, and MA$^\dagger$ models are generally worse than those of the same methods pretrained with in-domain medical image data---with the due caution that different backbones are used.
Also, the pretraining cannot consistently outperform training from scratch.
These may indicate that for transfer learning of medical images of modalities with significant domain gaps from natural images, pretraining on domain-specific data is critical and more effective than on large-scale natural image datasets.
}

Fig. \ref{fig:cmr_results} visualizes several representative examples corresponding to the results of different pretext tasks in Table \ref{tab:cmr}.
As we can see, our MTL produces segmentations closest to the ground truth, consistent with the quantitative evaluation. 

\begin{table}[!t]\color{black}
\centering
\caption{Classification performance (AUC) on the ACDC CMR dataset \citep{bernard2018deep}, and comparison of transfer learning performance with different pretraining methods: training from scratch (None), slice ordering \citep[Slice ord.;][]{zhang2017self}, jigsaw puzzle \citep[Jig. puzz.;][]{puzzle2016unsupervised}, Models Genesis \citep[Mod. Gen.;][]{zhou2019models}, anatomical position prediction \citep[APP;][]{bai2019self}, SimCLR \citep{chen2020simple}, BYOL \citep{grill2020bootstrap}, masked autoencoder (MA) \citep{he2022masked}, our relative location regression (Rel. loc.), relative orientation regression (Rel. ori.), and multitask learning (MTL).
{\color{blue}$^\dagger$: directly fine-tuned from ImageNet pretrained models with different backbones from other methods.}}
\label{tab:cmr_cls}
 \setlength{\tabcolsep}{1.mm}
\begin{adjustbox}{width=\linewidth}
\begin{threeparttable}
\begin{tabular}{cccccc}
\hline
Pretrain                  & \multicolumn{5}{c}{No. subjects\tnote{a}}                                                   \\ \cline{2-6} 
method                    & 1 (1.56\%)     & 4 (6.25\%)     & 16 (25\%)      & 32 (50\%)      & 64 (100\%)     \\ \hline
None                      & 0.608          & 0.825          & 0.861          & 0.909          & 0.920          \\
\cdashline{1-6}Slice ord. & 0.750          & 0.884          & 0.912          & 0.934          & 0.918          \\
Jig. puzz.                & 0.835          & 0.881          & 0.929          & 0.904          & 0.968          \\
Mod. Gen.                 & 0.768          & 0.893          & 0.937          & 0.928          & 0.928          \\
APP                       & 0.686          & 0.815          & 0.922          & 0.921          & 0.956          \\
\cdashline{1-6}SimCLR     & 0.567          & 0.821          & 0.921          & 0.924          & 0.925          \\
BYOL                      & 0.454          & 0.754          & 0.912          & 0.943          & 0.943          \\
MA                       & 0.598          & 0.868          & 0.931          & 0.937          & 0.935          \\
\cdashline{1-6}Rel. loc.  & 0.734          & 0.880          & 0.925          & 0.927          & 0.962          \\
Rel. ori.                 & 0.795          & 0.868          & 0.903          & 0.925          & 0.934          \\
MTL                       & \textbf{0.839} & \textbf{0.903} & \textbf{0.959} & \textbf{0.959} & \textbf{0.972} \\ 
\cline{1-6}{\color{blue} SimCLR$^\dagger$}     & {\color{blue} 0.484}          & {\color{blue} 0.717}          & {\color{blue} 0.841}          & {\color{blue} 0.866}          & {\color{blue} 0.873}          \\
{\color{blue} BYOL$^\dagger$}                      & {\color{blue} 0.392}          & {\color{blue} 0.735}          & {\color{blue} 0.893}          & {\color{blue} 0.925}          & {\color{blue} 0.935}          \\
{\color{blue} MA$^\dagger$}                       & {\color{blue} 0.489}          & {\color{blue} 0.811}          & {\color{blue} 0.818}          & {\color{blue} 0.827}          & {\color{blue} 0.840}          \\
\hline
\end{tabular}
\begin{tablenotes}\small
    \item[a] The number of subjects used for fine-tuning on the task of CMR semantic segmentation, with percentage with respect to the entire training dataset in parentheses.
    \end{tablenotes}
\end{threeparttable}
\end{adjustbox}
\end{table}

\begin{table*}[!t]
\centering
\caption{Results of multi-structural segmentation of SAX CMR images of the ACDC dataset \citep{bernard2018deep} ($\uparrow$: higher is better; $\downarrow$: lower is better), and comparison of transfer learning performance with different pretraining methods: training from scratch (None), slice ordering \citep[Slice ord.;][]{zhang2017self}, jigsaw puzzle \citep[Jig. puzz.;][]{puzzle2016unsupervised}, Models Genesis \citep[Mod. Gen.;][]{zhou2019models}, anatomical position prediction \citep[APP;][]{bai2019self}, {\color{black}SimCLR \citep{chen2020simple}, BYOL \citep{grill2020bootstrap}, masked autoencoder (MA) \citep{he2022masked},} our relative location regression (Rel. loc.), relative orientation regression (Rel. ori.), and multitask learning (MTL).
{\color{blue}$^\dagger$: directly fine-tuned from ImageNet pretrained models with different backbones from other methods.}
Format: mean (standard deviation).}
\label{tab:cmr}

\setlength{\tabcolsep}{1.mm}
\begin{adjustbox}{width=1.\linewidth}
\begin{tabular}{cccccccccc}
\hline
Amount & Pretrain &\multicolumn{2}{c}{Mean}  &\multicolumn{2}{c}{LV} &\multicolumn{2}{c}{RV} &\multicolumn{2}{c}{Myocardium} \\

\cline{3-4} \cline{5-10}

of data\textsuperscript{a} & method   &Dice $\uparrow$  & ASSD {(mm)} $\downarrow$  & Dice  $\uparrow$  & ASSD {(mm)} $\downarrow$  & Dice  $\uparrow$  & ASSD {(mm)} $\downarrow$  & Dice  $\uparrow$  & ASSD {(mm)} $\downarrow$  \\

\hline
\multirow{8}{*}{\makecell*[c]{1 subject\\(1.56\%):\\20 slices}}
    & None
    & 0.328 (0.286) & 20.223 (28.729)
    & 0.456 (0.338) & 27.463 (39.943)
    & 0.227 (0.216) & 21.054 (21.337)
    & 0.299 (0.238) & 12.151 (17.525)   \\ \cdashline{2-10}

    & Slice ord.
    & 0.437 (0.318) & 20.425 (28.849)
    & 0.507 (0.357) & 19.558 (29.433)
    & 0.341 (0.288) & 24.468 (28.677)
    & 0.462 (0.278) & 17.249 (27.946)   \\

    & Jig. puzz.
    & 0.410 (0.286) & 21.778 (29.537)
    & 0.486 (0.297) & 25.064 (35.956)
    & 0.370 (0.266) & 18.757 (20.099)
    & 0.372 (0.279) & 21.513 (30.008)    \\

    & Mod. Gen.
    & 0.470 (0.275) & 14.713 (22.634)
    & 0.569 (0.294) & 14.676 (24.579)
    & 0.374 (0.237) & 16.884 (23.920)
    & 0.466 (0.254) & 12.578 (18.742)    \\

    & APP
    & 0.461 (0.290) & 17.151 (31.190)
    & 0.547 (0.323) & 20.156 (39.117)
    & 0.375 (0.265) & 17.364 (23.287)
    & 0.461 (0.249) & 13.932 (28.748)  \\ \cdashline{2-10}

{\color{black} }                        & {\color{black} SimCLR} & {\color{black} 0.278 (0.288)} & {\color{black} 33.262 (44.645)}  & {\color{black} 0.328 (0.344)} & {\color{black} 33.503 (47.469)}  & {\color{black} 0.264 (0.249)} & {\color{black} 31.741 (40.985)}  & {\color{black} 0.243 (0.255)} & {\color{black} 33.262 (44.645)}  \\
{\color{black} }                        & {\color{black} BYOL}   & {\color{black} 0.230 (0.252)}  & {\color{black} 33.929 (43.196)}  & {\color{black} 0.232 (0.293)} & {\color{black} 33.140 (43.360)}  & {\color{black} 0.242 (0.216)} & {\color{black} 28.948 (38.385)}  & {\color{black} 0.216 (0.241)} & {\color{black} 39.699 (46.749)}  \\
{\color{black} }                        & {\color{black} MA}    & {\color{black} 0.326 (0.282)} & {\color{black} 26.841 (42.818)}  & {\color{black} 0.414 (0.336)} & {\color{black} 18.023 (30.621)}  & {\color{black} 0.256 (0.236)} & {\color{black} 40.991 (53.739)}  & {\color{black} 0.307 (0.239)} & {\color{black} 21.506 (42.817)}  \\ \cdashline{2-10}

    & Rel. loc.
    & 0.433 (0.306) & 18.333 (25.977)
    & 0.481 (0.360) & 20.638 (30.416)
    & 0.380 (0.266) & 17.671 (20.710)
    & 0.440 (0.275) & 16.691 (25.729)  \\

    & Rel. ori.
    & 0.471 (0.297) & 14.258 (24.692)
    & 0.567 (0.336) & \textbf{10.486} (18.307)
    & 0.394 (0.253) & \textbf{13.876} (16.098)
    & 0.453 (0.269) & 18.412 (34.688)  \\

    & MTL
    & \textbf{0.511} (0.289) & \textbf{13.900} (29.809)
    & \textbf{0.619} (0.317) & 14.665 (33.594)
    & \textbf{0.438} (0.241) & 17.034 (35.445)
    & \textbf{0.475} (0.270) & \textbf{10.001} (15.974) \\  
\hline

\multirow{8}{*}{\makecell*[c]{4 subjects\\(6.25\%):\\80 slices}}
    & None
    & 0.702 (0.228) & 4.754 (7.705)
    & 0.761 (0.258) & 5.138 (8.782)
    & 0.631 (0.241) & 6.506 (9.409)
    & 0.715 (0.151) & 2.618 (2.157)  \\ \cdashline{2-10}

    & Slice ord.
    & 0.745 (0.205) & 3.316 (4.150)
    & 0.826 (0.185) & 2.791 (4.512)
    & 0.659 (0.230) & 4.918 (4.420)
    & 0.750 (0.156) & 2.241 (2.788)  \\

    & Jig. puzz.
    & 0.783 (0.156) & 2.273 (3.290)
    & 0.856 (0.164) & 1.995 (4.794)
    & 0.721 (0.152) & 3.503 (2.491)
    & 0.771 (0.115) & 1.323 (0.891)  \\

    & Mod.Gen.
    & 0.798 (0.138) & 2.117 (2.095)
    & 0.875 (0.108) & 1.635 (2.287)
    & 0.736 (0.159) & 3.015 (1.950)
    & 0.782 (0.101) & 1.701 (1.710)  \\

    & APP
    & 0.766 (0.175) & 2.528 (2.768)
    & 0.828 (0.153) & 2.372 (2.727)
    & 0.711 (0.219) & 3.552 (3.416)
    & 0.758 (0.115) & 1.659 (1.431)  \\ \cdashline{2-10}

{\color{black} }                        & {\color{black} SimCLR} & {\color{black} 0.750 (0.192)}         & {\color{black} 3.348 (5.569)}                      & {\color{black} 0.802 (0.243)}         & {\color{black} 4.145 (8.937)}                      & {\color{black} 0.698 (0.179)}         & {\color{black} 3.946 (2.786)}  & {\color{black} 0.749 (0.115)}         & {\color{black} 1.953 (1.575)}  \\
{\color{black} }                        & {\color{black} BYOL}   & {\color{black} 0.762 (0.161)}        & {\color{black} 2.592 (2.382)}                      & {\color{black} 0.801 (0.184)}         & {\color{black} 3.062 (3.105)}                      & {\color{black} 0.764 (0.121)}         & {\color{black} 2.873 (1.939)}  & {\color{black} 0.722 (0.163)}         & {\color{black} 1.841 (1.661)}  \\
{\color{black} }                        & {\color{black} MA}    & {\color{black} 0.788 (0.167)}        & {\color{black} 2.259 (3.091)}                      & {\color{black} 0.842 (0.175)}         & {\color{black} 2.429 (4.516)}                      & {\color{black} 0.732 (0.197)}         & {\color{black} 2.944 (2.492)}  & {\color{black} 0.789 (0.089)}         & {\color{black} 1.404 (0.907)}  \\ \cdashline{2-10}

    & Rel. loc.
    & 0.794 (0.151) & 2.585 (2.915)
    & \textbf{0.882} (0.080) & \textbf{1.556} (1.350)
    & 0.733 (0.188) & 3.431 (3.102)
    & 0.766 (0.118) & 2.766 (3.499)  \\

    & Rel. ori.
    & 0.813 (0.136) & 2.226 (4.959)
    & 0.871 (0.154) & 2.639 (8.266)
    & \textbf{0.784} (0.139) & \textbf{2.562} (1.932)
    & 0.789 (0.083) & 1.476 (0.946)  \\

    & MTL
    & \textbf{0.814} (0.118) & \textbf{1.917} (1.668)
    & 0.869 (0.107) & 1.853 (1.967)
    & 0.777 (0.132) & 2.628 (1.721)
    & \textbf{0.795} (0.091) & \textbf{1.271} (0.770) \\
\hline
\multirow{8}{*}{\makecell*[c]{16 subjects\\(25\%):\\318 slices}}

    & None
    & 0.815 (0.129) & 2.064 (1.943)
    & 0.849 (0.135) & 2.326 (2.511)
    & 0.785 (0.144) & 2.611 (1.813)
    & 0.809 (0.093) & 1.255 (0.845)  \\ \cdashline{2-10}

    & Slice ord.
    & 0.813 (0.133) & 2.070 (2.121)
    & 0.837 (0.150) & 2.573 (2.966)
    & 0.794 (0.141) & 2.317 (1.670)
    & 0.806 (0.101) & 1.319 (1.016)  \\

    & Jig. puzz.
    & 0.815 (0.116) & 2.001 (1.869)
    & 0.845 (0.133) & 2.341 (2.581)
    & 0.809 (0.104) & 2.272 (1.502)
    & 0.790 (0.102) & 1.391 (1.001)  \\

    & Mod.Gen.
    & 0.819 (0.144) & 2.078 (2.263)
    & 0.856 (0.146) & 2.222 (2.935)
    & 0.792 (0.176) & 2.415 (2.149)
    & 0.810 (0.087) & 1.596 (1.325)  \\

    & APP
    & 0.815 (0.141) & 2.144 (2.676)
    & 0.847 (0.142) & 2.458 (3.102)
    & 0.785 (0.177) & 2.625 (3.030)
    & 0.810 (0.079) & 1.348 (1.311)  \\ \cdashline{2-10}

{\color{black} }                        & {\color{black} SimCLR} & {\color{black} 0.824 (0.129)}        & {\color{black} 1.992 (3.011)}                      & {\color{black} 0.852 (0.168)}         & {\color{black} 2.581 (4.877)}                      & {\color{black} 0.815 (0.112)}         & {\color{black} 1.969 (1.263)}  & {\color{black} 0.807 (0.093)}         & {\color{black} 1.429 (1.073)}  \\
{\color{black} }                        & {\color{black} BYOL}   & {\color{black} 0.823 (0.132)}        & {\color{black} 1.838 (2.791)}                      & {\color{black} 0.844 (0.165)}         & {\color{black} 2.301 (4.348)}                      & {\color{black} 0.825 (0.126)}         & {\color{black} 2.039 (1.799)}  & {\color{black} 0.799 (0.089)}         & {\color{black} 1.173 (0.726)}  \\
{\color{black} }                        & {\color{black} MA}    & {\color{black} 0.818 (0.158)}        & {\color{black} 2.031 (2.641)}                      & {\color{black} 0.861 (0.142)}         & {\color{black} 2.004 (2.724)}                      & {\color{black} 0.766 (0.205)}         & {\color{black} 2.947 (3.324)}  & {\color{black} \textbf{0.826} (0.086)}         & {\color{black} 1.143 (0.914)}  \\ \cdashline{2-10}

    & Rel. loc.
    & 0.831 (0.148) & 1.938 (3.411)
    & \textbf{0.873} (0.167) & 1.944 (4.873)
    & 0.794 (0.169) & 2.654 (2.981)
    & 0.826 (0.148) & 1.215 (1.187)  \\

    & Rel. ori.
    & 0.834 (0.115) & \textbf{1.599} (1.622)
    & 0.866 (0.112) & \textbf{1.683} (2.016)
    & \textbf{0.837} (0.135) & \textbf{1.782} (1.683)
    & 0.801 (0.084) & 1.331 (0.940)  \\

    & MTL
    & \textbf{0.840} (0.136) & 1.602 (2.699)
    & 0.868 (0.177) & 1.922 (4.263)
    & 0.825 (0.131) & 1.865 (1.661)
    & \textbf{0.826} (0.078) & \textbf{1.019} (0.641)  \\

\hline
\multirow{8}{*}{\makecell*[c]{32 subjects\\(50\%):\\618 slices}}

    & None
    & 0.851 (0.112) & 1.552 (2.779)
    & 0.887 (0.153) & 1.671 (4.508)
    & 0.824 (0.089) & 2.126 (1.343)
    & 0.844 (0.061) & 0.858 (0.465)  \\ \cdashline{2-10}

    & Slice ord.
    & 0.875 (0.107) & 1.400 (2.569)
    & 0.880 (0.160) & 1.893 (4.142)
    & 0.877 (0.076) & 1.515 (1.317)
    & 0.867 (0.053) & 0.792 (0.531)  \\

    & Jig. puzz.
    & 0.880 (0.104) & 1.187 (2.568)
    & 0.905 (0.152) & 1.351 (4.241)
    & 0.859 (0.078) & 1.565 (1.118)
    & 0.875 (0.046) & 0.645 (0.296)  \\

    & Mod.Gen.
    & 0.884 (0.066) & 1.075 (1.079)
    & 0.912 (0.072) & 1.095 (1.468)
    & 0.875 (0.058) & 1.427 (0.986)
    & 0.866 (0.058) & 0.702 (0.324)  \\

    & APP
    & 0.871 (0.095) & 1.339 (1.616)
    & 0.913 (0.088) & 1.046 (1.739)
    & 0.832 (0.114) & 2.232 (1.846)
    & 0.869 (0.055) & 0.738 (0.406)  \\ \cdashline{2-10}

{\color{black} }                        & {\color{black} SimCLR} & {\color{black} 0.869 (0.108)}        & {\color{black} 1.430 (2.364)}                      & {\color{black} 0.884 (0.156)}         & {\color{black} 1.766 (3.861)}                      & {\color{black} 0.857 (0.089)}         & {\color{black} 1.695 (1.170)}  & {\color{black} 0.867 (0.052)}         & {\color{black} 0.830 (0.539)}  \\
{\color{black} }                        & {\color{black} BYOL}   & {\color{black} 0.897 (0.062)}        & {\color{black} 0.873 (0.866)}                      & {\color{black} \textbf{0.929} (0.047)}         & {\color{black} \textbf{0.687} (0.775)}                      & {\color{black} 0.885 (0.067)}         & {\color{black} 1.259 (1.124)}  & {\color{black} 0.877 (0.056)}         & {\color{black} 0.672 (0.404)}  \\
{\color{black} }                        & {\color{black} MA}    & {\color{black} 0.886 (0.071)}        & {\color{black} 1.089 (1.183)}                      & {\color{black} 0.915 (0.080)}         & {\color{black} 0.997 (1.482)}                      & {\color{black} 0.864 (0.076)}         & {\color{black} 1.622 (1.179)}  & {\color{black} 0.879 (0.040)}         & {\color{black} 0.647 (0.353)}  \\ \cdashline{2-10}

    & Rel. loc.
    & 0.884 (0.075) & 1.149 (1.289)
    & 0.913 (0.082) & 1.070 (1.640)
    & 0.865 (0.082) & 1.671 (1.288)
    & 0.874 (0.049) & 0.705 (0.405)  \\

    & Rel. ori.
    & 0.885 (0.073) & 1.087 (1.286)
    & 0.908 (0.085) & 1.034 (1.630)
    & 0.872 (0.079) & 1.546 (1.336)
    & 0.875 (0.043) & 0.680 (0.376)  \\

    & MTL
    & \textbf{0.900} (0.055) & \textbf{0.850} (0.815)
    & {0.926} (0.051) & 0.735 (0.945)
    & \textbf{0.891} (0.058) & \textbf{1.177} (0.908)
    & \textbf{0.882} (0.047) & \textbf{0.639} (0.331) \\
\hline
\multirow{8}{*}{\makecell*[c]{64 subjects\\(100\%):\\1188 slices}}

    & None
    & 0.868 (0.081) & 1.302 (1.358)
    & 0.920 (0.049) & 0.727 (0.701)
    & 0.828 (0.106) & 2.317 (1.809)
    & 0.855 (0.041) & 0.863 (0.467)  \\ \cdashline{2-10}

    & Slice ord.
    & 0.887 (0.067) & 1.069 (1.150)
    & 0.920 (0.077) & 1.003 (1.613)
    & 0.879 (0.058) & 1.376 (0.909)
    & 0.862 (0.050) & 0.828 (0.617)  \\

    & Jig. puzz.
    & 0.886 (0.067) & 1.037 (1.116)
    & 0.913 (0.072) & 1.021 (1.504)
    & 0.877 (0.071) & 1.418 (1.054)
    & 0.867 (0.045) & 0.671 (0.298)  \\

    & Mod.Gen.
    & 0.898 (0.061) & 0.883 (0.962)
    & 0.915 (0.073) & 0.866 (1.177)
    & 0.893 (0.067) & 1.197 (1.058)
    & 0.887 (0.033) & 0.586 (0.290)  \\

    & APP
    & 0.903 (0.056) & 0.799 (0.908)
    & 0.930 (0.067) & 0.714 (1.253)
    & 0.891 (0.050) & 1.127 (0.806)
    & 0.889 (0.036) & 0.558 (0.281)  \\ \cdashline{2-10}

{\color{black} }                        & {\color{black} SimCLR} & {\color{black} 0.897 (0.054)}        & {\color{black} 0.921 (0.878)}                      & {\color{black} 0.928 (0.053)}         & {\color{black} 0.790 (1.047)}                      & {\color{black} 0.882 (0.048)}         & {\color{black} 1.306 (0.925)}  & {\color{black} 0.882 (0.046)}         & {\color{black} 0.668 (0.389)}  \\
{\color{black} }                        & {\color{black} BYOL}   & {\color{black} 0.901 (0.055)}        & {\color{black} 0.841 (0.813)}                      & {\color{black} 0.925 (0.059)}         & {\color{black} 0.843 (1.044)}                      & {\color{black} \textbf{0.897} (0.048)}         & {\color{black} \textbf{1.048} (0.811)}  & {\color{black} 0.882 (0.048)}         & {\color{black} 0.632 (0.379)}  \\
{\color{black} }                        & {\color{black} MA}    & {\color{black} 0.902 (0.058)}        & {\color{black} 0.833 (0.798)}                      & {\color{black} 0.935 (0.053)}         & {\color{black} 0.584 (0.723)}                      & {\color{black} 0.884 (0.065)}         & {\color{black} 1.334 (0.964)}  & {\color{black} 0.886 (0.039)}         & {\color{black} 0.583 (0.289)}  \\ 

\cdashline{2-10}

    & Rel. loc.
    & 0.907 (0.054) & 0.793 (0.769)
    & 0.939 (0.040) & 0.570 (0.679)
    & 0.896 (0.060) & 1.167 (0.934)
    & 0.886 (0.046) & 0.641 (0.480)  \\

    & Rel. ori.
    & 0.906 (0.053) & 0.789 (0.882)
    & 0.934 (0.047) & 0.634 (0.821)
    & 0.894 (0.061) & 1.190 (1.168)
    & 0.890 (0.038) & 0.544 (0.228)  \\  

    & MTL
    & \textbf{0.910} (0.058) & \textbf{0.719} (0.932)
    & \textbf{0.941} (0.038) & \textbf{0.526} (0.528)
    & \textbf{0.897} (0.075) & 1.119 (1.425)
    & \textbf{0.893} (0.042) & \textbf{0.510} (0.233)  \\

\cline{2-10}
& {\color{blue} SimCLR$^\dagger$} & \color{blue}0.884 (0.113) & \color{blue}1.159 (2.867) & \color{blue}0.903 (0.153) & \color{blue}1.149 (4.608) & \color{blue}0.877 (0.109) & \color{blue}1.319 (1.708) & \color{blue}0.871 (0.052) & \color{blue}0.661 (0.113) \\
& {\color{blue} BYOL$^\dagger$} & \color{blue}0.872 (0.077) & \color{blue}1.202 (1.216) & \color{blue}0.883 (0.091) & \color{blue}1.484 (1.447) & \color{blue}0.881 (0.082) & \color{blue}1.286 (1.374) & \color{blue}0.851 (0.046) & \color{blue}0.836 (0.486) \\
& {\color{blue} MA$^\dagger$} & \color{blue}0.885 (0.069) & \color{blue}1.035 (1.154) & \color{blue}0.912 (0.075) & \color{blue}1.083 (1.529) & \color{blue}0.887 (0.066) & \color{blue}1.216 (1.133) & \color{blue}0.855 (0.052) & \color{blue}0.807 (0.538) \\
    
\hline
\multicolumn{10}{l}{\textsuperscript{a} Amount of data used for fine-tuning on the target task; format: number of subjects (percentage with respect to the entire training dataset): number of slices.}
\end{tabular}
\end{adjustbox}
\end{table*}

{\color{black}Table \ref{tab:cmr_cls} shows the AUCs for diagnosing previous myocardial infarction, dilated cardiomyopathy, hypertrophic cardiomyopathy, abnormal right ventricle, and normal cases on the ACDC dataset.
Almost all the compared pretext tasks improve performance upon the train-from-scratch baseline with all different amounts of fine-tuning data for semantic segmentation (except for the contrastive learning based approaches in few-subject settings).
Meanwhile, our MTL pretraining achieves the highest AUCs for all amounts of fine-tuning data.
These results display similar trends to Table \ref{tab:cmr}, which is expected as the classifier depends on the segmentation results, and the segmentation model fine-tuned from our MTL pretraining performs the best overall.
}

\subsubsection{Knee MRI Diagnosis}

Table \ref{tab:knee} presents the mean AUCs for diagnosing abnormality, ACL tear, and meniscal tear based on sagittal T2W fat suppressed knee MRI.
Above all, it is noted that for training from scratch, increasing the number of subjects from 80 to 400 brings no substantial change in the mean AUC (0.433, 0.429 and 0.474 for 80, 200, and 400 subjects, respectively), and further increasing to 800 subjects improves the AUC marginally to 0.562, suggesting the challenge of the downstream task.
Second, although consistently better performance is achieved using transfer learning than training from scratch---as expected, it is to our surprise that fine-tuning ImageNet pretrained models with other pretext tasks actually diminishes the transfer learning efficacy of the ImageNet pretraining in most settings.
In contrast, our MTL substantially improves upon the ImageNet pretraining when the number of subjects is small (by $\sim$0.07 and $\sim$0.05 with 80 and 200 subjects, respectively), and maintains comparable performance to the ImageNet pretraining when the number of subjects is 400 or larger.
In fact, with 80 and 200 subjects, our relative location or orientation regression task alone leads to substantial improvement upon the ImageNet pretraining.
Lastly, in contrast with the downstream task of CMR segmentation, the pretext task of relative location regression consistently yields superior performance to the relative orientation regression on knee MRI diagnosis.

\begin{table}[t]
\centering
\caption{Classification performance (mean AUC) on sagittal T2W fat-saturated knee MRI of the Stanford dataset \citep{MRnet}, and comparison of transfer learning performance with different pretraining methods: training from scratch (None), ImageNet \citep{russakovsky2015imagenet}, Models Genesis \citep[Mod. Gen.;][]{zhou2019models}, anatomical position prediction \citep[APP;][]{bai2019self}, jigsaw puzzle \citep[Jig. puzz.;][]{puzzle2016unsupervised}, slice ordering \citep[Slice ord.;][]{zhang2017self}, {\color{black}masked autoencoder (MA) \citep{he2022masked}, BYOL \citep{grill2020bootstrap}, SimCLR \citep{chen2020simple},} our relative location regression (Rel. loc.), relative orientation regression (Rel. ori.), and multitask learning (MTL).
{\color{blue}$^\dagger$: directly fine-tuned from ImageNet pretrained models with different backbones from other methods.}}
\label{tab:knee}
\begin{adjustbox}{width=\linewidth}
\begin{threeparttable}
\begin{tabular}{ccccc}
\hline
Pretrain                                     & \multicolumn{4}{c}{No. subjects\tnote{a}}                                                                                         \\ \cline{2-5} 
method                                       & 80 (10\%)                    & 200 (25\%)                   & 400 (50\%)                   & 800 (100\%)                  \\ \hline
None                                         & 0.433                        & 0.429                        & 0.474                        & 0.562                        \\
ImageNet                                     & 0.682                        & 0.737                        & {\ul 0.799}                  & \textbf{0.824}               \\
\cdashline{1-5}
Mod. Gen.                                    & 0.631                        & 0.722                        & 0.707                        & 0.699                        \\
APP                                          & 0.654                        & 0.713                        & 0.719                        & 0.723                        \\
Jig. puzz.                    & 0.632                        & 0.703                        & 0.770                        & 0.793                        \\
Slice ord.                                   & 0.625                        & 0.740                        & 0.773                        & 0.792                        \\
\cdashline{1-5}
{\color{black} MA}                   & {\color{black} 0.548} & {\color{black} 0.583} & {\color{black} 0.618} & {\color{black} 0.624} \\
{\color{black} BYOL}                  & {\color{black} 0.577} & {\color{black} 0.641} & {\color{black} 0.690}  & {\color{black} 0.753} \\
{\color{black} SimCLR} & {\color{black} 0.698} & {\color{black} 0.761} & {\color{black} 0.774} & {\color{black} 0.800} \\
\cdashline{1-5}Rel. loc.                     & {\ul 0.740}                  & {\ul 0.781}                  & 0.791                        & 0.815                        \\
Rel. ori.                                    & 0.703                        & 0.774                        & 0.787                        & 0.803                        \\
MTL                                          & \textbf{0.755}               & \textbf{0.783}               & \textbf{0.801}               & {\ul 0.819}                  \\ 
\cline{1-5}
{\color{blue} MA$^\dagger$}                   & {\color{blue} 0.421} & {\color{blue} 0.462} & {\color{blue} 0.519} & {\color{blue} 0.536} \\
{\color{blue} BYOL$^\dagger$}                  & {\color{blue} 0.440} & {\color{blue} 0.531} & {\color{blue} 0.533}  & {\color{blue} 0.552} \\
{\color{blue} SimCLR$^\dagger$} & {\color{blue} 0.432} & {\color{blue} 0.474} & {\color{blue} 0.496} & {\color{blue} 0.530} \\
\hline
\end{tabular}
\begin{tablenotes}\small
    \item[a] The number of subjects used for fine-tuning on the target task, with percentage with respect to the entire training dataset in parentheses.
    \end{tablenotes}
\end{threeparttable}
\end{adjustbox}
\end{table}

\section{Discussion and conclusion}\label{sec:conclude_discuss}
In this work, we proposed two complementary
self-supervising pretext tasks for transfer learning of medical image data with anatomy-oriented imaging planes: regressing relative orientations and locations of a set of views.
Both pretext tasks were based on the spatial relationship among the views and were simple and straightforward.
We conducted thorough experiments with two representative types of such medical image data (cardiac and knee MRI) and two typical downstream tasks (semantic segmentation and computer-aided diagnosis) in medical image analysis.
The results showed that the proposed pretext tasks were not only feasible for pretraining (i.e., with tractable learnability), but also effective in transfer learning for the typical downstream tasks.
In addition, our proposed pretext tasks led to superior performance on the downstream tasks to existing self-supervising pretext tasks.

Both of our proposed pretext tasks were conceptually simple and easy to implement. Their learnability was validated in Section \ref{sec:exp:perf_pre}.
On the one hand, all the evaluated metrics were quite reasonable for the task of relative location regression on both datasets, as presented in Table \ref{tab:loc}.
In addition, after combining with the task of relative orientation regression, all the evaluated metrics were improved.
This suggested that the other task provided positive feedbacks for predicting the slice locations.
On the other hand, the models learned to predict stable intersecting lines for the task of relative orientation regression on both datasets, as shown in Figs. \ref{fig:heatmap_CMR} and \ref{fig:heatmap_knee}, across different slice locations, cardiac phases (CMR only), and pathological status.
These results suggested that the models had acquired knowledge about the crucial landmarks that defined the set of imaging planes, robust against varying conditions.

We demonstrated transfer learning with the proposed pretext tasks on two typical downstream tasks in medical image analysis: semantic segmentation and classification, and on two representative datasets: cardiac and knee MRI.
The experimental results showed that the proposed pretext tasks could effectively improve downstream task performance upon training from scratch, especially in the low-data regime.
{\color{purple}We conjecture that the performance gain could result from (1) robust representations learned from the pretext tasks \citep{tian2020rethinking} and (2) the regularizing effect of the pretraining against overfitting \citep{gidaris2019boosting}.}
When combing both pretext tasks together, the multitask learning achieved further improvements upon pretraining with either of them,
suggesting that the two tasks could boost each other 
without conflict and thus were truly complementary.
Compared to the other self-supervising pretext tasks, ours yielded better performance on the downstream tasks, indicating superiority in transfer learning for medical image data with anatomy-oriented imaging planes.
Particularly, when used to fine-tune the ImageNet pretrained models,
our pretext tasks can further improve the transfer learning performance in low-data regime.
This is desirable when ImageNet pretrained models are available.
{\color{black}We attributed the improvements to the smaller distances between our pretext tasks and potential downstream tasks for the targeted data group.
More specifically, the proposed pretext tasks were intensively focused on learning the anatomy of the structure of interest being imaged, which was helpful to target tasks on that specific structure.
%
In addition, understanding the images as a whole in the pretext tasks might also help downstream tasks that require whole images as input, compared with patch-based pretraining.
}
In the future, we plan to also apply our pretext tasks to other popular tasks in medical image analysis, e.g., lesion detection.

When employed individually,
relative orientation regression was more effective than relative location regression on the downstream task of CMR segmentation, whereas the opposite was observed on the downstream task of knee MRI classification.
We conjecture that this might be a manifestation of the different distances between the pretext and downstream tasks, i.e., relative orientation regression was closer to semantic segmentation than relative location regression, and the opposite for classification.
{\color{black}Meanwhile, in prescribing knee MRI, the axial and coronal planes were (roughly) perpendicular and parallel to the middle line of the femur and tibia in the sagittal views, respectively (Fig. \ref{fig:heatmap_knee}). 
This may have simplified the pretext task of relative orientation regression for knee MRI since locating the knee cap and the generally upright femur and tibia might be easier than prescribing CMR (although we implemented random rotation as part of the online data augmentation), thus reducing its difficulty and efficacy in pretraining the networks.}

We notice that the pairwise slice ordering pretext task proposed by \citet{zhang2017self} generally deteriorated the transfer performance of the ImageNet pretraining (Table \ref{tab:knee}).
This was because the network had difficulty in ordering similar slices symmetric about a center.
In contrast, with the center-symmetric mapping (Eqn. (\ref{eq:sym_loc})), our relative location regression substantially improves upon the ImageNet pretraining.
{\color{black}In this work, we employ the sine function to map similar slices in symmetrical cases to similar regression values. 
There is a potential to let the network learn a more general mapping from the image data, like the relation network, which learned the distance function for low-shot recognition \citep{sung2018learning}.
}

A limitation of this work was that the effectiveness of the proposed self-supervising pretext tasks was only evaluated in the scenario of transfer learning, where the networks were pretrained on a different dataset before transferred to the target dataset for the downstream task.
This was because we had no access to a suitable dataset that had both spatial information properly recorded for pretraining and annotations available for fine-tuning and evaluation.
Despite the potential domain gap \citep{tsai2018learning} in our experimental setting, transfer learning with our pretext tasks turned out effective.
In practice, it is very likely that the dataset for the target task can also be used for self-supervised pretraining with the pretext tasks, {\color{purple}given the spatial information is properly recorded for valid volumetric analysis}.
In such practical scenarios, we expect the impact of our pretext tasks to be more significant considering the elimination of the domain gap.\footnote{\color{black}After this work was completed, we became aware of a concurrent work that published a dataset suitable for pretraining by our proposed pretext tasks and evaluation on downstream tasks \citep{martin2023deep}.
We hope to evaluate our method on this candidate dataset in the future and also encourage the research community to do so with our published code.}

{\color{black}Also, we note that our proposed pretext tasks were exclusive to multi-view medical image data with anatomy-oriented imaging planes.
Yet, this data group comprises a significant subset of clinical imaging data, e.g., MRI of various organs and body parts and the standard mammography views.
Given the abundance of such data in clinics, we believe our methodology contributed significantly to the medical image analysis community by highlighting the underexplored potential of exploiting anatomy position information in medical image pretraining.
}

{\color{blue}Lastly, this work was focused on effective network pretraining by self-supervised learning for medical image data with anatomy-oriented imaging planes.
Accordingly, we employed straightforward network and training configurations for fine-tuning on target tasks to emphasize the effect of pretraining.
We expect more advanced deep pipelines for medical image analysis, e.g., the nnU-Net \citep{isensee2021nnu}, would benefit from incorporating weights pretrained by our proposed pretext tasks on applicable data.
}

\section*{Acknowledgments}
The authors gratefully acknowledge the support of the National Natural Science Foundation of China under Key Program 62236009, General Program 61876032 (to S.G.), Shenzhen Science and Technology Program under JCYJ20210324140807019 (to S.G.)

\bibliographystyle{model2-names.bst}\biboptions{authoryear}
\bibliography{refs}
\end{document}